\documentclass[10pt,twocolumn,letterpaper]{article}
\usepackage[table]{xcolor} 
\usepackage{cvpr}

\usepackage{times}
\usepackage{textcomp}
\usepackage{paralist}
\usepackage{footmisc}
\usepackage{placeins}
\usepackage{balance}

\usepackage{graphicx}
\graphicspath{{data/}}

\usepackage[labelformat=simple]{subcaption}
\DeclareCaptionLabelSeparator{periodspace}{.\quad}
\usepackage{amsmath, amssymb, mathrsfs}
\usepackage{mathtools}
\usepackage{amsthm}

\DeclareMathOperator*{\argmax}{\arg\!\max}
\DeclareMathOperator*{\argmin}{\arg\!\min}
\newcommand{\vect}[1]{{\mathbf{#1}}}
\newcommand{\mat}[1]{{\mathbf{#1}}}
\newcommand{\set}[1]{{\mathcal{#1}}}
\newcommand{\norm}[2]{\left\| #1 \right\|_{#2}}
\newcommand{\transpose}[1]{#1^\mathrm{T}}

\newcommand{\Real}{{\mathbb R}}

\usepackage[algo2e,ruled,vlined,linesnumbered]{algorithm2e}
\SetKwComment{Comment}{//\ }{}
\makeatletter
\newcommand{\nosemic}{\renewcommand{\@endalgocfline}{\relax}}
\newcommand{\dosemic}{\renewcommand{\@endalgocfline}{\algocf@endline}}
\makeatother

\usepackage{booktabs}
\usepackage{multirow}
\usepackage{tabu}
\usepackage{rotating}
\usepackage{siunitx}
\usepackage{threeparttable}

\newcommand{\firstcolor}{\cellcolor{black!15}}
\newcommand{\secondcolor}{\cellcolor{black!7}}

\usepackage[pagebackref=true,breaklinks=true,colorlinks,bookmarks=false]{hyperref}

\usepackage[capitalise,compress]{cleveref}
\crefname{section}{Sec.}{secs.}
\crefname{figure}{Fig.}{figs.}
\crefname{table}{Tab.}{tabs.}
\crefname{equation}{Eq.}{eqs.}
\crefrangelabelformat{equation}{(#3#1#4--#5#2#6)}
\crefname{problem}{Problem}{problems}
\creflabelformat{problem}{(#2#1#3)}
\crefname{algorithm}{Alg.}{algs.}
\creflabelformat{algorithm}{#2#1#3}

\cvprfinalcopy 
\def\cvprPaperID{755} 

\begin{document}

\title{Nonnegative Matrix \emph{Underapproximation} for Robust Multiple Model Fitting\thanks{Work partially supported by NSF, ARO, ONR, and NGA.}}

\author{
\shortstack{%
	Mariano Tepper\thanks{M.~Tepper performed almost all of this work while at Duke University.}\\
	Flatiron Institute, Simons Foundation\\
	{\tt\small mtepper@flatironinstitute.org}
}%
\hspace{2cm}%
\shortstack{%
	Guillermo Sapiro\\
	ECE, Duke University\\
	{\tt\small guillermo.sapiro@duke.edu}
}%
}

\maketitle

\begin{abstract}
   In this work, we introduce a highly efficient algorithm to address the nonnegative matrix underapproximation (NMU) problem, i.e., nonnegative matrix factorization (NMF) with an additional underapproximation constraint. NMU results are interesting as, compared to traditional NMF, they present additional sparsity and part-based behavior, explaining unique data features. To show these features in practice, we first present an application to the analysis of climate data.
   We then present an NMU-based algorithm to robustly fit multiple parametric models to a dataset. The proposed approach delivers state-of-the-art results for the estimation of multiple fundamental matrices and homographies, outperforming other alternatives in the literature and exemplifying the use of efficient NMU computations.
\end{abstract}

\section{Introduction}

Nonnegative Matrix Factorization (NMF)~\cite{Paatero1994} has been successfully applied as a data analysis technique, mainly because its part-based representation and sparsity. Unfortunately, the problem of finding a rank-$r$ NMF of a nonnegative matrix $\mat{A}$ is NP-hard~\cite{Vavasis2010}.
However, a rank-one NMF can be obtained in polynomial time.
On the one hand, the rank-one singular value decomposition (SVD) of $\mat{A}$ is an optimal solution to the problem
\begin{equation}
	\min_{\operatorname{rank}(\mat{Z}) = 1} \norm{\mat{A} - \mat{Z}}{F}^2 .
\end{equation}
On the other hand, the Perron-Frobenius theorem implies that the SVD factors with the largest singular value of a nonnegative matrix are themselves nonnegative~\cite[Chapter 8]{Meyer2000}. Hence, the SVD provides an optimal rank-one NMF.

Following a standard approach in the literature~\cite{papalexakis2013,witten2009ssvd} we might be tempted to iterate two steps until the desired rank is reached:
\begin{compactenum}
	\item compute the rank-one SVD $s \vect{u} \transpose{\vect{v}}$ of $\mat{A}$, and
	\item set $\mat{A} = \mat{A} - s \vect{u} \transpose{\vect{v}}$.
\end{compactenum}
However, the matrix $( \mat{A} - s \vect{u} \transpose{\vect{v}} )$ is not necessarily nonnegative, presenting a serious obstacle to the overall approach.

Nonnegative Matrix \emph{Underapproximation} (NMU)~\cite{Gillis2010,Gillis2011,Gillis2013} appears as a solution to this challenge by posing the related problem
\begin{equation}
	\min_{\substack{ \vect{u} \in \Real^{m}, \vect{v} \in \Real^{n}}} \norm{\mat{A} - \vect{u} \transpose{\vect{v}} }{F}^2
	\quad \text{s.t.} \quad
	\begin{gathered}
	\vect{u}, \vect{v} \geq \vect{0} , \\
	\mat{A} \geq \vect{u} \transpose{\vect{v}}.
	\end{gathered}
	\tag{NMU}	
\label[problem]{eq:nmu}
\end{equation}
Now, $\mat{A} - \vect{u} \transpose{\vect{v}}$ is constrained to be a nonnegative matrix, rendering the previous iterative approach possible.

\begin{figure}[t]
	\includegraphics[width=\columnwidth]{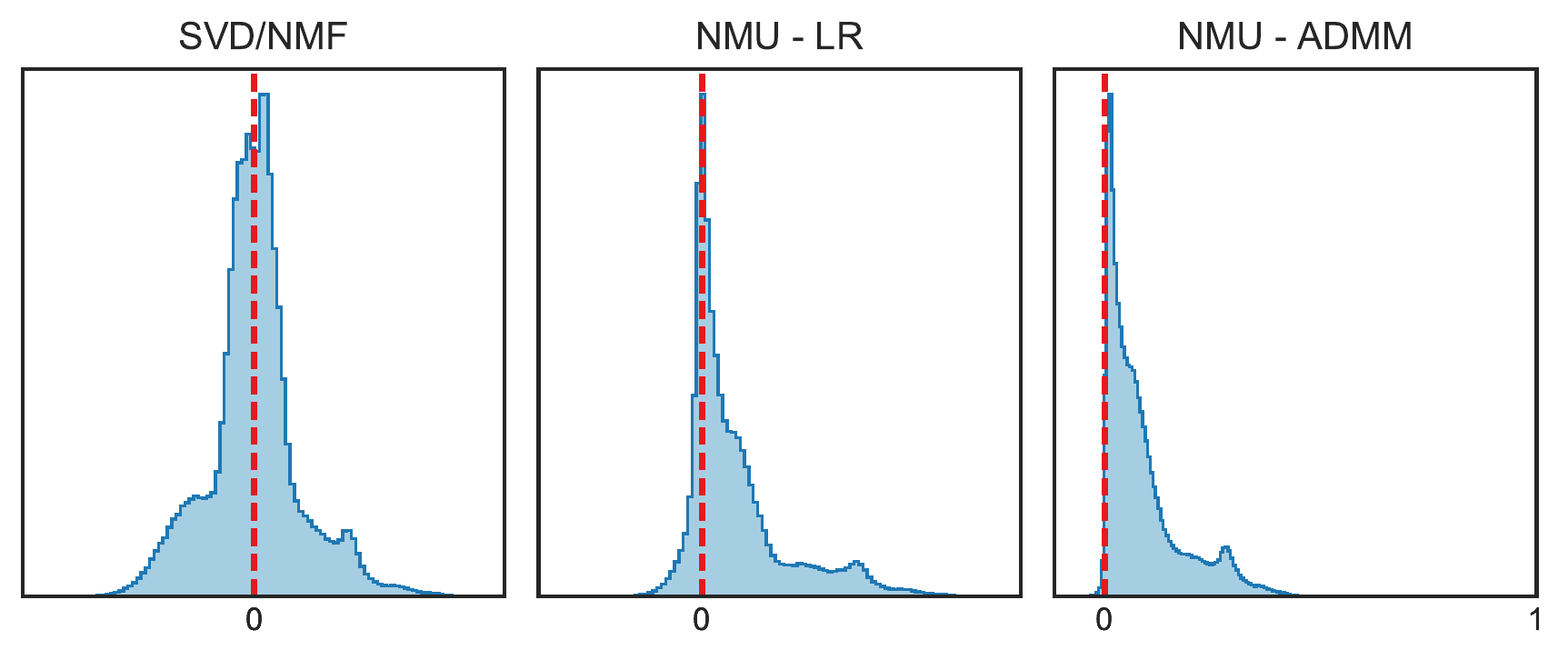}
	
	\caption{In this example we show the histogram of the values of $\mat{A} - \vect{u} \transpose{\vect{v}}$. NMF (which coincides with the SVD in the rank-one case) is not expected to yield any particular pattern. NMU must yield nonnegative values. Clearly, the Lagrangian relaxation method (LR), reported as the state-of-the-art in the literature until now, does not correctly enforce the underapproximation constraint while the here proposed ADMM method does. Refer to \cref{fig:climate} to see the dataset details.}
	\label{fig:nmu_comparison}
\end{figure}

In this work, we first present a new efficient NMU technique, based on the Alternating Direction Method of Multipliers (ADMM).\footnote{\label{note:software}Source code available at \url{https://goo.gl/xSqKQ4}.}
The resulting algorithm is, to the best of our knowledge, the only NMU algorithm in the literature to efficiently solve \cref{eq:nmu} while enforcing all its constraints (see \cref{fig:nmu_comparison}). Additionally, it is easy to implement and delivers high-quality results.

We then present an application of the proposed NMU technique to the problem of robustly fitting multiple parametric models in a given dataset.\footref{note:software}
Fitting multiple instances of a given parametric model to data corrupted by noise and outliers is a widespread problem in computer vision. It is encountered in a diverse set of applications such as finding lines/circles/ellipses in images, homography estimation in stereo vision, motion estimation/segmentation, and analysis of 3D point clouds.
We draw the connection between NMU and this important problem in computer vision and pattern recognition and show that the proposed formulation leads (1) to NMU factors that are naturally interpretable and (2) to state-of-the-art results. This is also the case for an example on climate data here provided.

\noindent\textbf{Organization.}
The remainder of this paper is organized as follows. In \cref{sec:nmu} we present the proposed NMU technique  and provide an example pattern recognition application. In \cref{sec:mpme} we introduce the robust fitting formulation. We present computer vision experimental results in \cref{sec:results} and finally provide some closing remarks in Section~\ref{sec:conclusions}.

\noindent\textbf{Notation.}
Let $\mat{X}$ be a matrix. $(\mat{X})_{ij}$, $(\mat{X})_{:j}$, $(\mat{X})_{i:}$ denote the $(i,j)$-th entry of $\mat{X}$, the $j$-th column of $\mat{X}$, and the $i$-th row of $\mat{X}$, respectively.

\section{Nonnegative Matrix Underapproximation}
\label{sec:nmu}

In several practical scenarios, the state-of-the-art Lagrangian relaxation~\cite{Gillis2010,Gillis2011,Gillis2013} did not yield good results. In particular, this algorithm delivers factors $\vect{u}, \vect{v}$ that violate the constraint $\mat{A} \geq \vect{u} \transpose{\vect{v}}$, defeating their own purpose and usefulness.
We show such an example in \cref{fig:nmu_comparison}.

In this work, we propose to use ADMM to solve \cref{eq:nmu}.
In short, ADMM solves convex optimization problems by breaking them into smaller subproblems, which are individually easier to handle. It has also been extended to handle non-convex problems, e.g., to solve several flavors of NMF~\cite{Fevotte2011,Tepper2014consensus,tepper2016compressed,xu2012nmf}.

\cref{eq:nmu} can be equivalently re-formulated as
\begin{equation}
	\min_{\substack{ \vect{u} \in \Real^{m}, \vect{v} \in \Real^{n}\\ \mat{R} \in \Real^{m \times n}}} \tfrac{1}{2} \norm{\mat{R}}{F}^2 \quad \text{s.t.} \quad
	\begin{gathered}[t]
		\mat{R} = \mat{A} - \vect{u} \transpose{\vect{v}} , \\
		\vect{u}, \vect{v} \geq \vect{0} , \mat{R} \geq \vect{0} ,
	\end{gathered}
	\label[problem]{eq:nmuEquiv}
\end{equation}
and we consider its augmented Lagrangian,
\begin{align}
	\mathscr{L} (\vect{u}, \vect{v}, \mat{R}, \mat{\Gamma})
	=& \tfrac{1}{2} \norm{\mat{R}}{F}^2 
	+ \mat{\Gamma} \bullet (\mat{A}- \vect{u} \transpose{\vect{v}} - \mat{R}) \nonumber \\
&+ \tfrac{\gamma}{2} \norm{\mat{A} - \vect{u} \transpose{\vect{v}} - \mat{R}}{F}^2 ,
\end{align}
where $\mat{\Gamma} \in \Real^{m \times n}$ is a Lagrange multiplier, $\gamma$ is a penalty parameter, and $\mat{B} \bullet \mat{C} = \sum_{i,j} (\mat{B})_{ij} (\mat{C})_{ij}$ for matrices $\mat{B}, \mat{C}$ of the same dimensions.

The ADMM algorithm works in a coordinate descent fashion, successively minimizing $\mathscr{L}$ with respect to $\vect{u}, \vect{v}, \mat{R}$, one at a time while fixing the others at their most recent values, i.e.,
\begin{subequations}
	\begin{align}
	\vect{u}_{k} &= \argmin_{\vect{u} \geq \vect{0}} \mathscr{L} (\vect{u}, \vect{v}_{k-1}, \mat{R}_{k-1}, \mat{\Gamma}_{k-1}) , \\
	\vect{v}_{k} &= \argmin_{\vect{v} \geq \vect{0}} \mathscr{L} (\vect{u}_{k}, \vect{v}, \mat{R}_{k-1}, \mat{\Gamma}_{k-1}) , \\
	\mat{R}_{k} &= \argmin_{\mat{R} \geq \mat{0}} \mathscr{L} (\vect{u}_{k}, \vect{v}_{k}, \mat{R}, \mat{\Gamma}_{k-1}) ,
	\end{align}
\end{subequations}
and then updating the multiplier $\mat{\Gamma}_{k}$.
For the problem at hand, each of these steps can be written in closed form as
\begin{subequations}
	\begin{align}
	&\vect{u}_{k} = \mathscr{P}_+ \left( \mat{M} \transpose{\vect{v}_{k-1}} / \left( \transpose{\vect{v}_{k-1}} \vect{v}_{k-1} \right) \right) ,\\ 
	&\vect{v}_{k} = \mathscr{P}_+ \left( \transpose{\vect{u}_{k}} \mat{M} / \left( \transpose{\vect{u}_{k}} \vect{u}_{k} \right) \right) ,\\
	&\mat{R}_{k} = \mathscr{P}_+ \left( \tfrac{1}{1 + \gamma} \left( \gamma \left( \mat{A} - \vect{u}_{k} \transpose{\vect{v}_{k}} \right) + \mat{\Gamma}_{k-1} \right) \right) , \\
	&\mat{\Gamma}_{k} = \mat{\Gamma}_{k-1} + \xi \gamma \left( \mat{A} - \vect{u}_{k} \transpose{\vect{v}_{k}} - \mat{R}_{k} \right) ,
	\end{align}
	\label{eq:admm_nmf}
\end{subequations}
where $\mat{M} = \mat{A} - \mat{R}_{k} + \gamma^{-1} \mat{\Gamma}_{k-1}$ and, for any matrix $\mat{B}$,
	$(\mathscr{P}_+ (\mat{B}))_{ij} = \max\left\{ (\mat{B})_{ij}, 0 \right\}$.
In practice, we set $\gamma = 1$ and $\xi = 1$. We name NMU-ADMM the iterations in \cref{eq:admm_nmf}.

\begin{figure}[t]
	\centering
	\begin{tabu} to \columnwidth {@{\hspace{0pt}} X[2,c,m] @{\hspace{4pt}}  X[8,c,m] @{\hspace{0pt}}}
	\includegraphics[width=\linewidth]{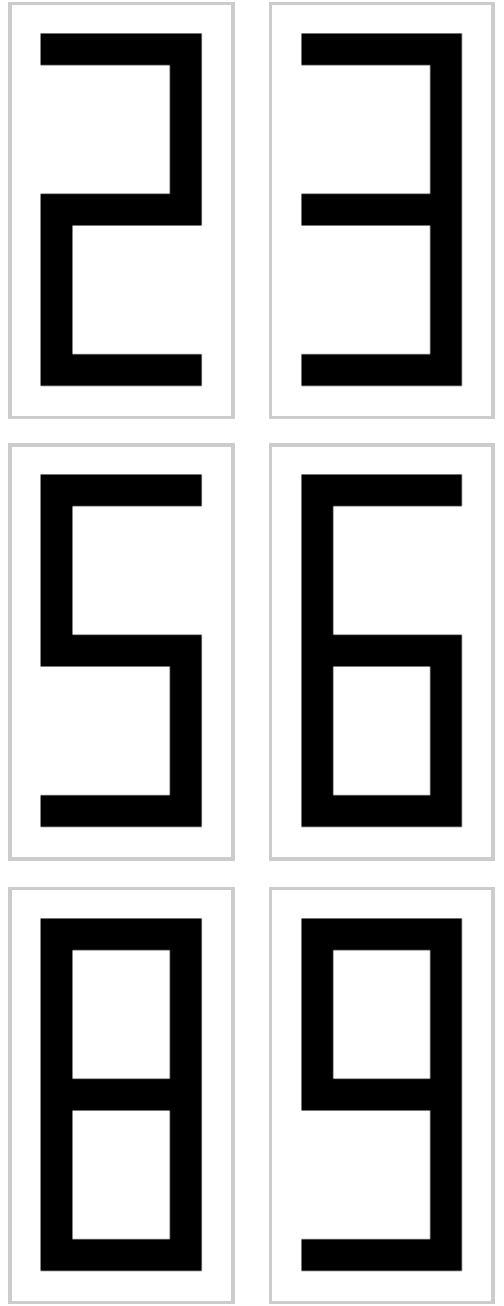} &
	\shortstack{
		\includegraphics[width=\linewidth]{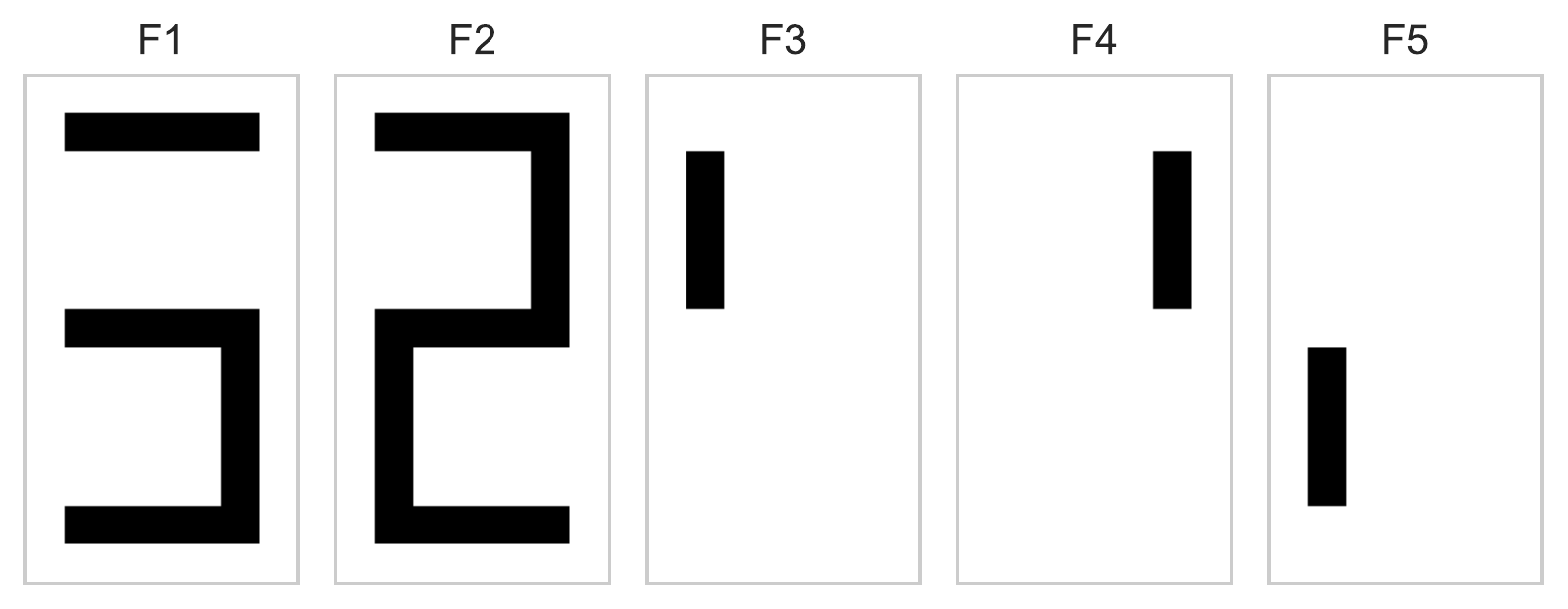}\\
		\includegraphics[width=\linewidth]{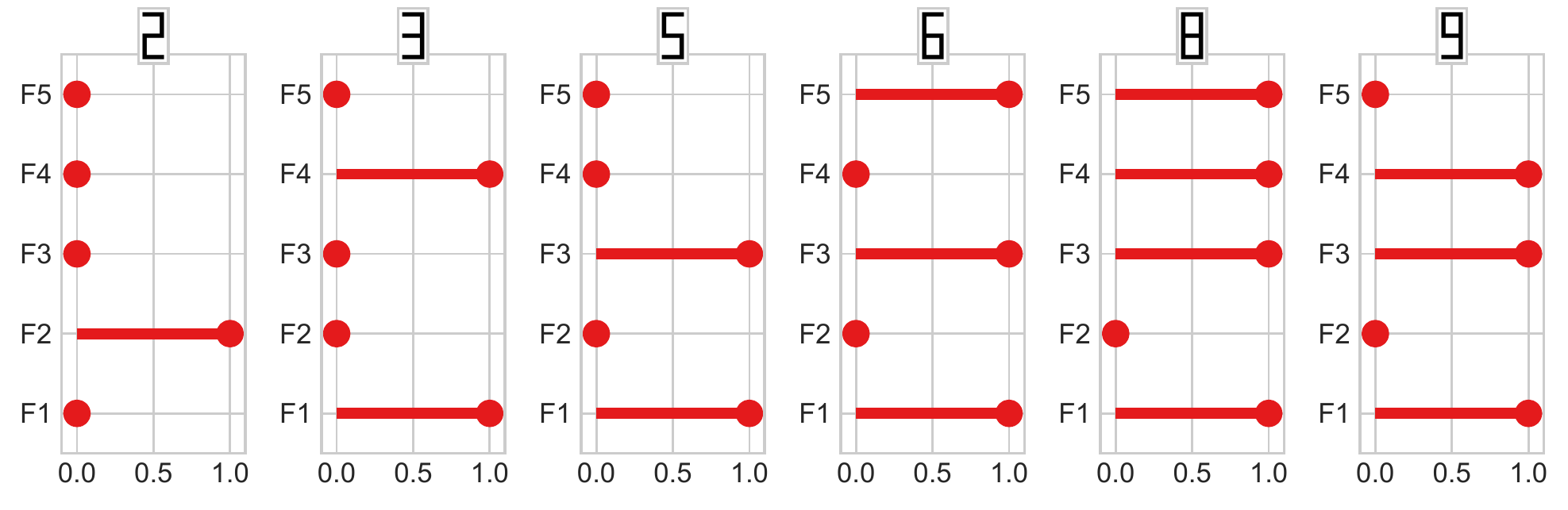}
	}
	\end{tabu}
	
	\caption{\textbf{Left:} Six binary images, that were reshaped as column vectors and concatenated to form the input matrix. The proposed NMU algorithm achieves perfect reconstruction with 5 factors. \textbf{Top right:} The obtained left factors: they are sparse and naturally provide a part-based representation. \textbf{Bottom right:} The values of the right factors for each reconstructed image; they can be interpreted as the contribution of each left factor to the reconstruction. Notice how these factors are also naturally sparse.}
	\label{fig:digits}
\end{figure}

%
%
%
%
%
%
%
%

\begin{figure*}[t]
	\centering

	\begin{footnotesize}
	\begin{minipage}{.2\textwidth}
		\includegraphics[width=.95\linewidth]{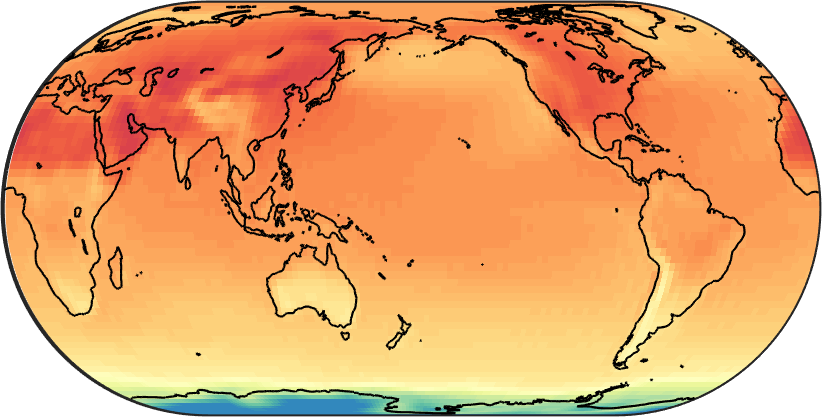} \\[6pt]
		\includegraphics[width=.95\linewidth]{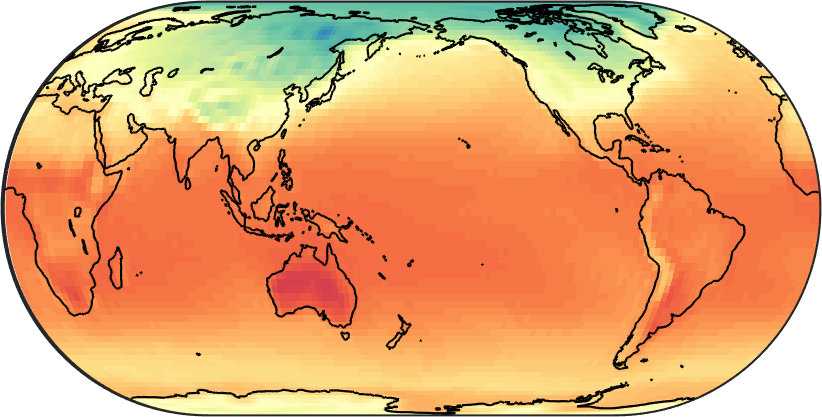} 	
	\end{minipage}%
	\begin{tabu} to .8\textwidth { @{\hspace{0cm}} X[c,m] *{3}{@{\hspace{.2cm}} X[c,m]} @{\hspace{0cm}} }
		\includegraphics[width=\linewidth]{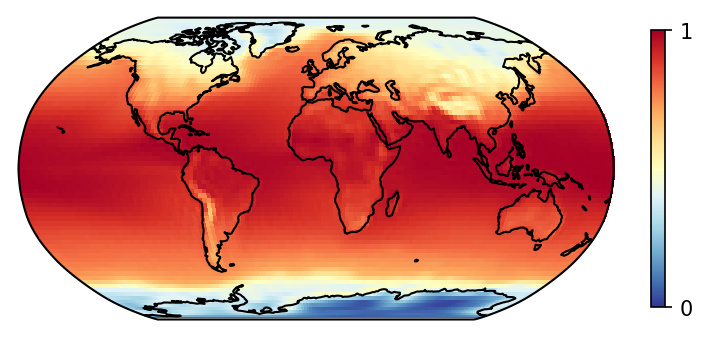} &
		\includegraphics[width=\linewidth]{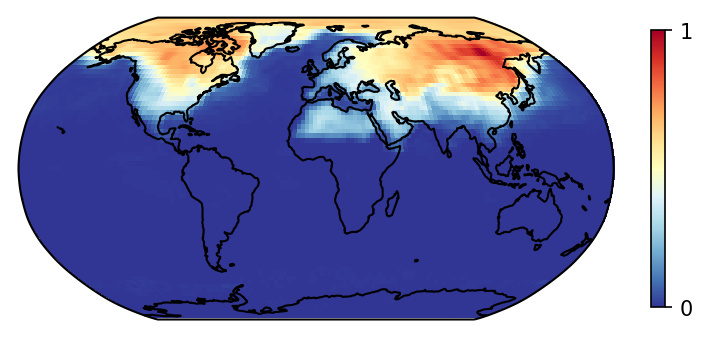} &
		\includegraphics[width=\linewidth]{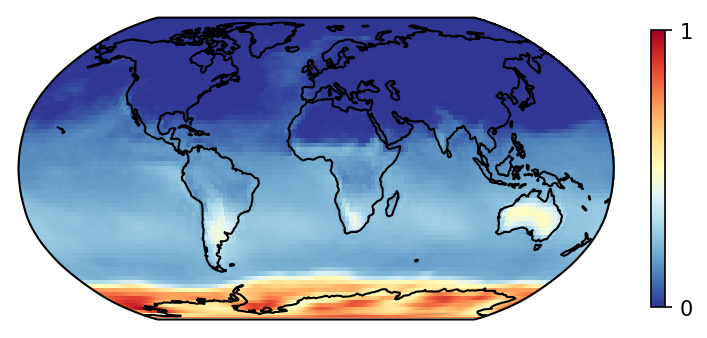} &
		\includegraphics[width=\linewidth]{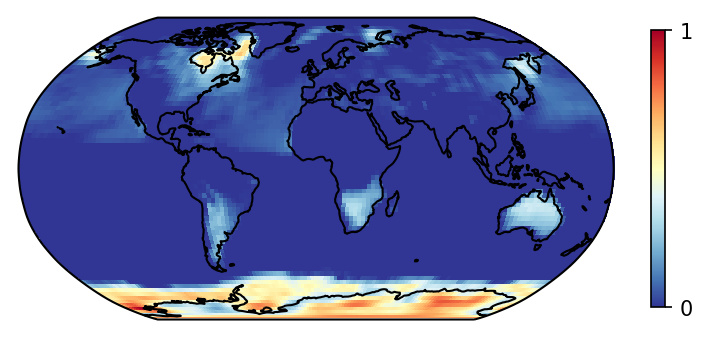} \\
		
		\\[-6pt]
		
		Energy: $96.84\%$ &
		Energy: $1.39\%$ &
		Energy: $1.17\%$ &
		Energy: $0.17\%$ \\

		\includegraphics[width=\linewidth]{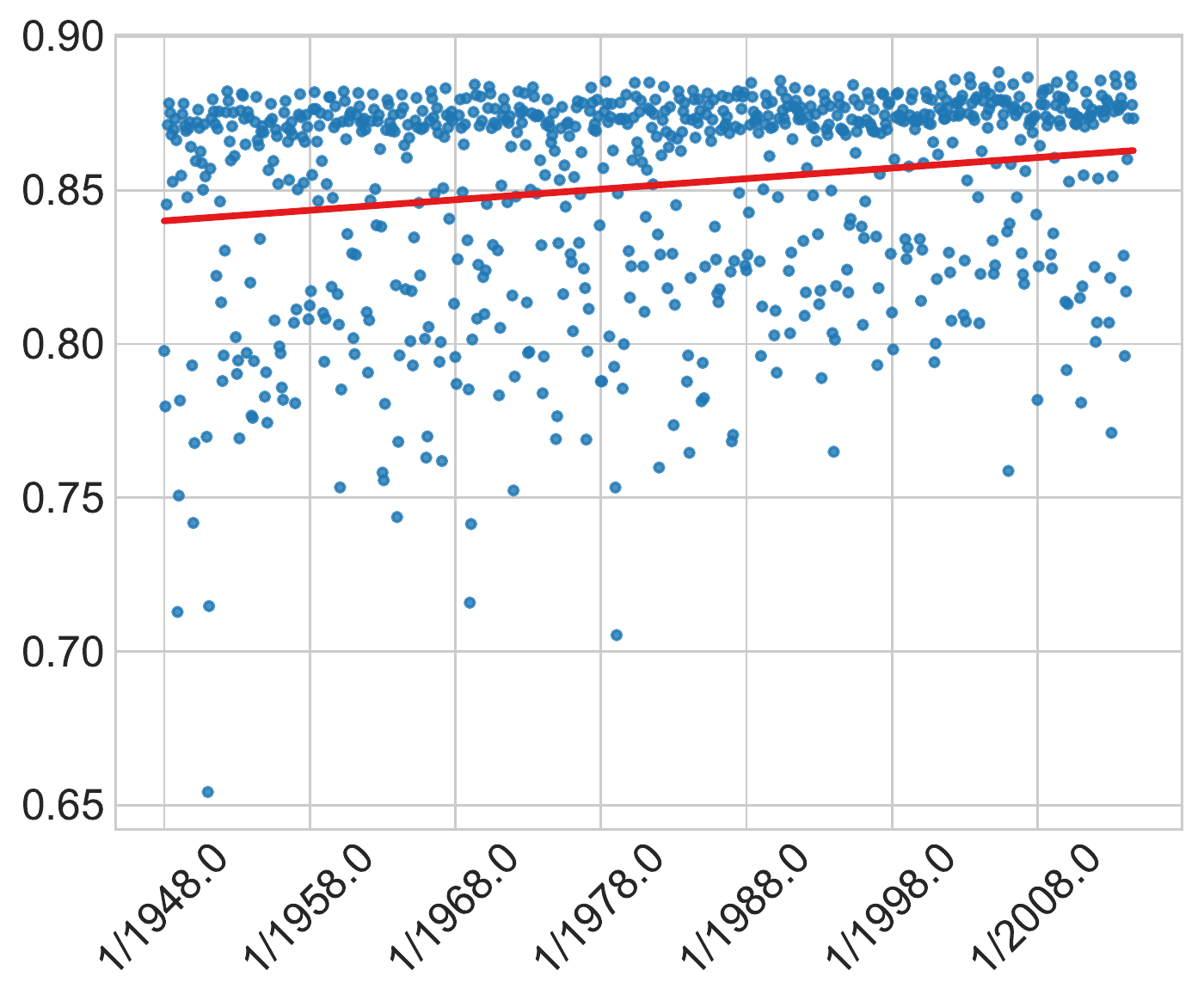} &
		\includegraphics[width=\linewidth]{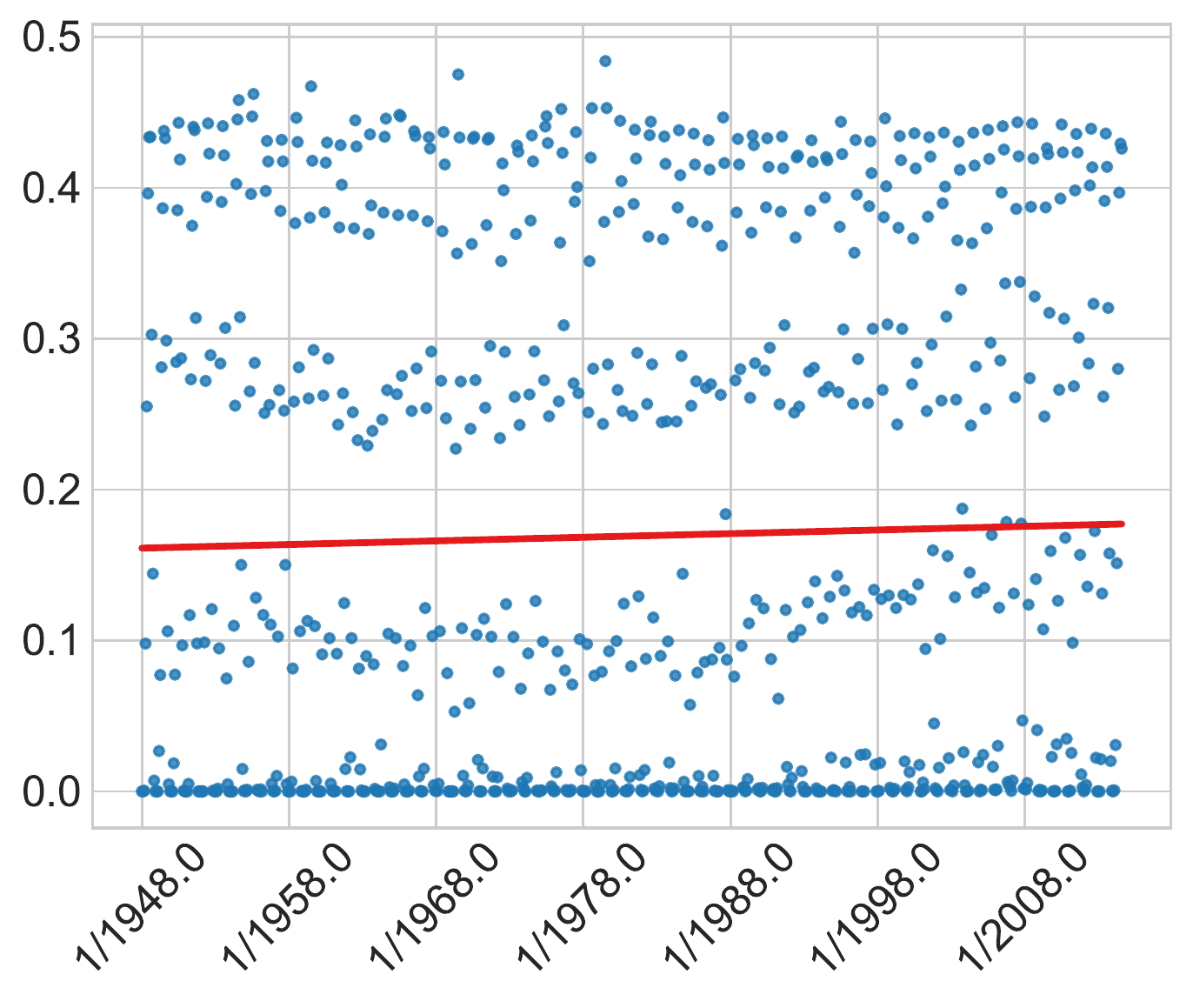} &
		\includegraphics[width=\linewidth]{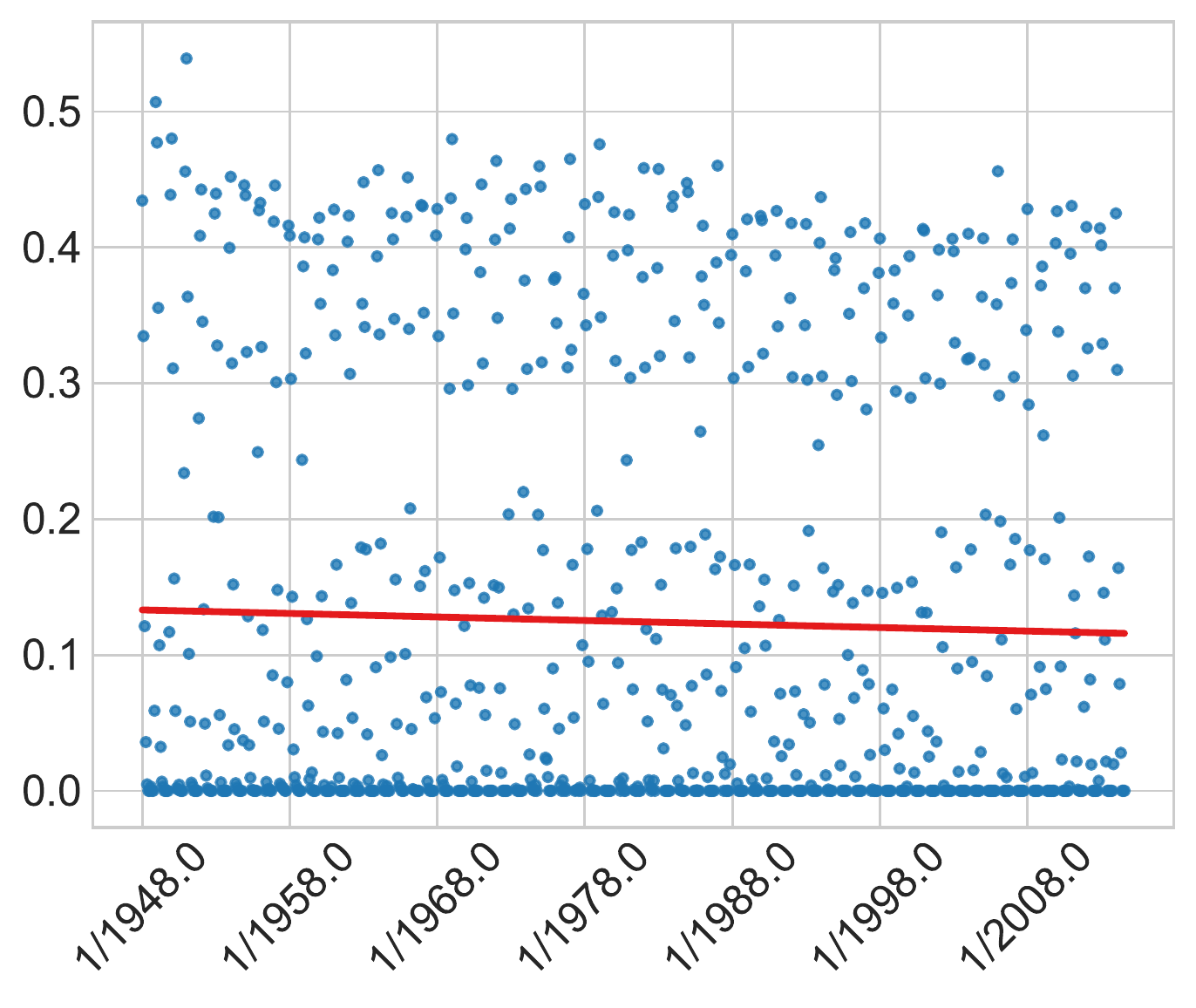} &
		\includegraphics[width=\linewidth]{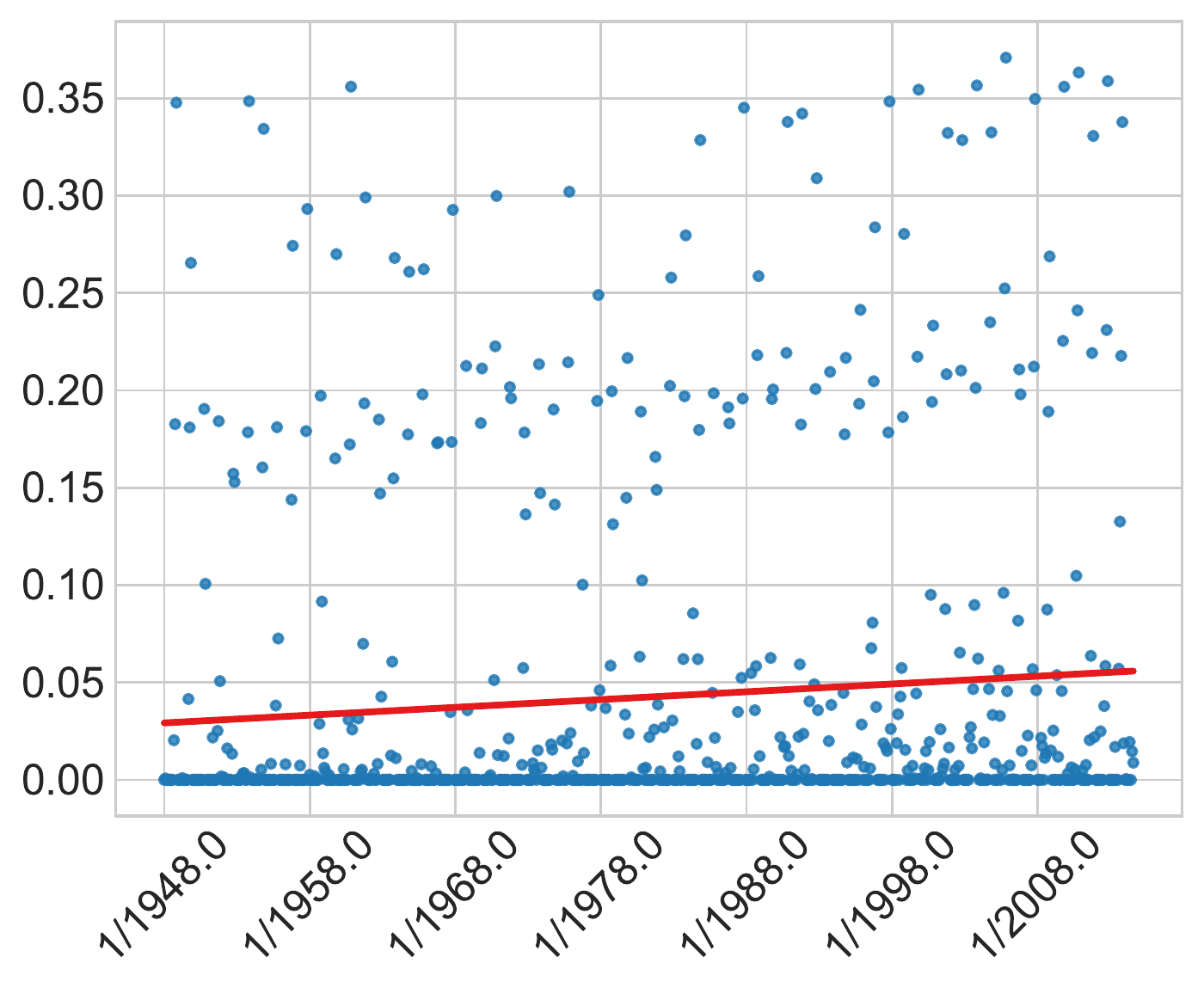} 
	\end{tabu}	
	\end{footnotesize}

	\caption{\textbf{NMU of climate data} (\url{http://www.esrl.noaa.gov/psd/repository/}). The data contains monthly mean surface temperatures arranged in a $144 \times 73$ grid since 1948 (800 months in total), forming a $10512 \times 800$ matrix. On the first column, we reproduce the NMF factors from~\cite{tepper2016compressed}, which represent the main seasonal effects (inverted winter and summer in the north and in the south). On the remaining columns, we show the first four NMU factors (left and right factors in the top and bottom rows, respectively), which amount to $98.64\%$ of the matrix energy. Their patterns are very different from NMF, first extracting a main global component and then providing part-based insights on different areas. It is also interesting to note that an upward trend can be perceived in the loadings (right factors) of the first two components, something that is not perceived with NMF, see~\cite{tepper2016compressed}.}
	\label{fig:climate}
\end{figure*}

As with most matrix factorization problems, there is a scaling indeterminacy in \cref{eq:nmu}, we can multiply and divide $\vect{u}$ and $\vect{v}$ respectively by a scalar constant and the result will remain unchanged. We thus arbitrarily set $\max (\vect{u}_k) = 1$ throughout the algorithm iterations and scale $\vect{v}_k$ accordingly. The algorithm terminates when the relative change in both $\vect{u}$ and $\vect{v}$ is below a given tolerance $\tau$.

\noindent\textbf{Initialization.}
It is natural to use the rank-one SVD to initialize $\vect{u}_0$ and $\vect{v}_0$. Let $\vect{x}, s, \vect{y}$ be the rank-one SVD of $\mat{A}$. Following our convention that $\max (\vect{u}_k) = 1$, we set
\begin{equation}
\vect{u}_0 = \vect{x} / \norm{\vect{x}}{\infty} ,
\quad\quad
\vect{v}_0 = \norm{\vect{x}}{\infty} s \, \vect{y} .
\end{equation}

\noindent\textbf{Extracting multiple NMU factors.}
As stated in the introduction, the algorithm for extracting multiple factors is straightforward. We iterate the following two steps:
\begin{compactenum}
	\item extract rank-one NMU factors $\vect{\hat{u}}, \vect{\hat{v}}$ from $\mat{A}$, and
	\item $\mat{A} \gets \mat{A} - \vect{\hat{u}} \transpose{\vect{\hat{v}}}$.
\end{compactenum}
A simple toy example illustrating the appealing qualities of NMU and of multi-factor NMU-ADMM is shown in \cref{fig:digits}.

\subsection{Illustrative application: Climate data analysis}

NMF provides a very rich descriptive power for climate datasets~\cite{tepper2016compressed}. The evidence of low rank models in climate data is of interest by itself, as they provide a concise way for describing the data. Nonnegativiy is a useful addition since, under this model, the effects of different factors cannot cancel each other, providing models that are easily interpretable. Furthermore, the rank of the approximation can be estimated soundly in an online fashion.

In the first example we show that NMU provides radically different insights, when compared with NMF.
The technical details and results of an experiment using climate data are shown in \cref{fig:climate}.
Analyzing the NMU factors, we see that we first obtain one main global component, with nonzero entries both in the left and right factors. These factors are sparse in the remaining components (up to numerical precision), even if sparsity is not formally promoted.

\begin{figure}[t]
	\centering
	\includegraphics[width=.85\columnwidth]{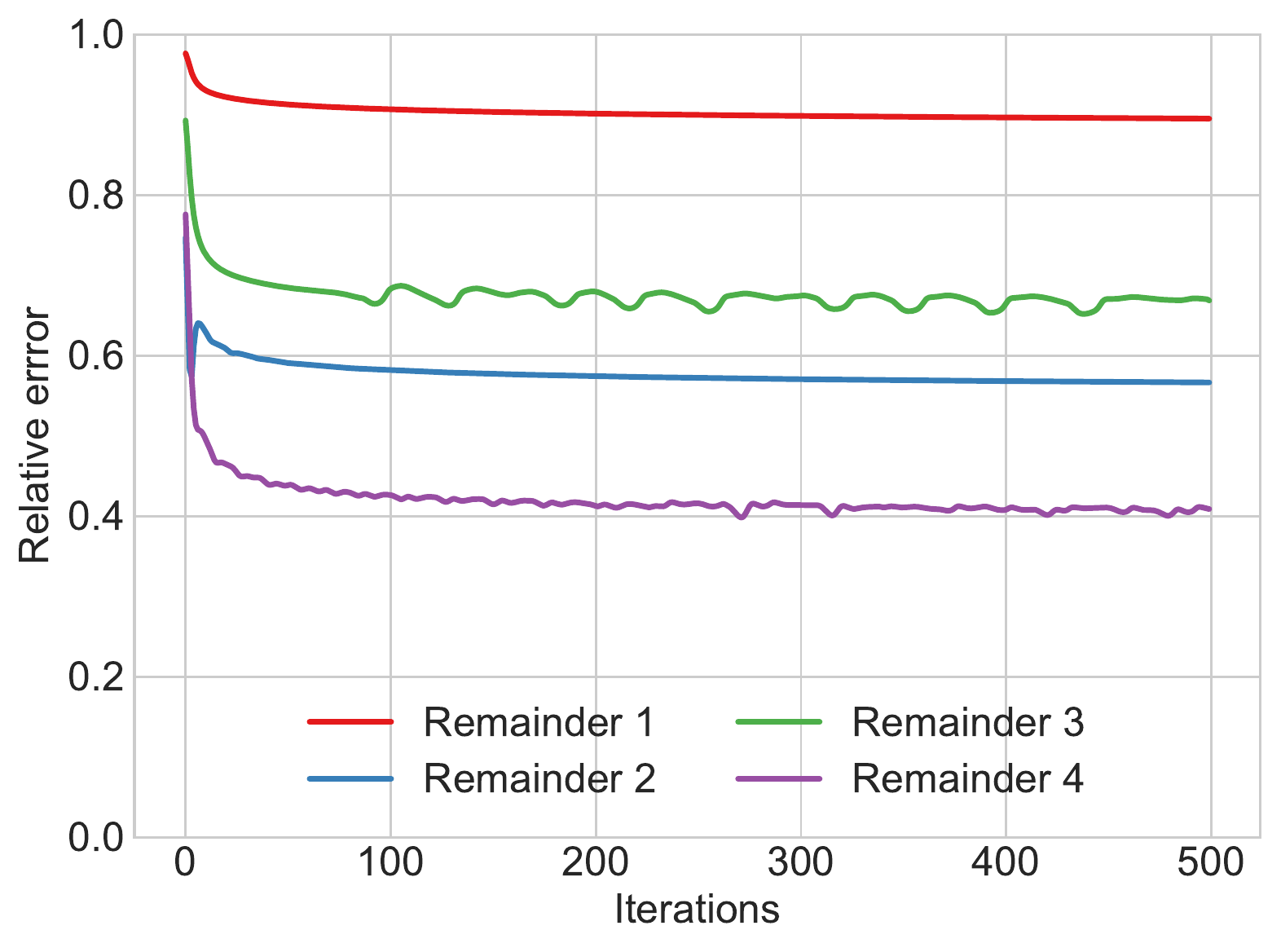}
	
	\caption{Convergence of $\norm{\mat{R}}{F} / \norm{\mat{A}}{F}$ in NMU-ADMM for the four factors computed in \cref{fig:climate} (recall that $\mat{A}$ is deflated after computing each NMU factor). In a few iterations, this value is already very close to the value after 500 iterations. Continuing the iterations improves the result, gaining marginal accuracy.}
	\label{fig:convergence}
\end{figure}

In \cref{fig:convergence}, we plot the error/remainder $\mat{R}$ for every NMU factor in \cref{fig:climate}. The algorithm converges very quickly to a steady solution. Let us point out that these are typical cases of the algorithm's performance.
A keen observer might object that the convergence is steadier for the first two factors than for the last two.
This can be explained by their high energy ($98.23\%$ of the total energy), making easier to extract coherent patterns (as these will be more evident).

\section{Robust model fitting}
\label{sec:mpme}

We now illustrate how to use NMU and in particular the proposed ADMM algorithm for solving multi-model estimation problems in computer vision.

Finding a single parametric model instance is a robust fitting problem that is hard on its own. When an \emph{unknown} number of instances might be present in the dataset the difficulty increases due to the unavoidable emergence of pseudo-outliers (data points that belong to one structure and are usually outliers to any other structure).
Thus, we face the problem of simultaneous robust estimation of model parameters and attribution of data points to the estimated models. These two problems are intrinsically intertwined.

The given dataset $\{ \vect{x}_i \}_{i=1}^{m}$ is composed of $m$ geometric objects to which we seek to fit one or more multiple instances of a parametric model.
Formally, a model $\mu$ is the zero level-set of a smooth parametric function $f_\mu (\vect{x}; \theta)$,
\begin{equation}
	\mu(\theta) = \{ \vect{x} \in \Real^d ,\ f_\mu (\vect{x}; \theta) = 0 \} ,
	\label{eq:model}
\end{equation}
where $\theta$ is a parameter vector.
The error associated with the datum $\vect{x} \in \set{X}$ with respect to the model $\mu(\theta)$ is
\begin{equation}
	\operatorname{e}_\mu (\vect{x}, \theta) = \min_{\vect{x}' \in \mu(\theta)} \operatorname{dist} (\vect{x}, \vect{x}') ,
	\label{eq:pointModelDistance}
\end{equation}
where $\operatorname{dist}$ is an appropriate distance function.
We also define the soft-membership function as
\begin{equation}
	\operatorname{sm}_\mu (\vect{x}, \theta) =
		\begin{cases}
		\exp \left( - d^2 / (2 \sigma^2) \right)
		& \text{if } d \leq 3 \sigma; \\
		0 & \text{otherwise,}
	\end{cases}
	\label{eq:soft_membership}
\end{equation}
where $d = \operatorname{e}_\mu (\vect{x}, \theta)$ and $\sigma$ is the main (and only) parameter of the proposed approach.

\begin{figure*}[t]
	\centering
	\begin{minipage}[b]{.33\textwidth}
		\centerline{
			\hfill%
			\includegraphics[width=.078\textwidth]{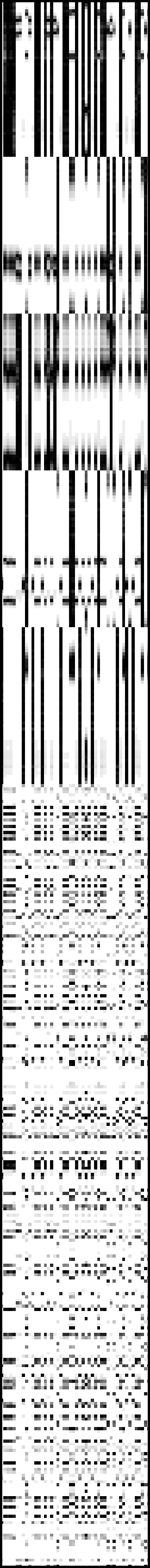}%
			\hfill%
			\includegraphics[width=.078\textwidth]{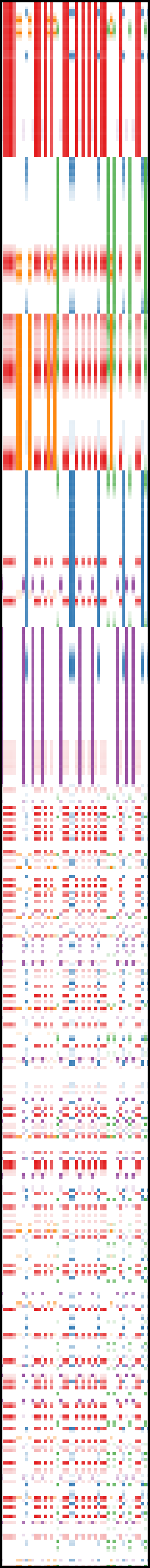}%
			\hfill%
			\includegraphics[width=.625\textwidth]{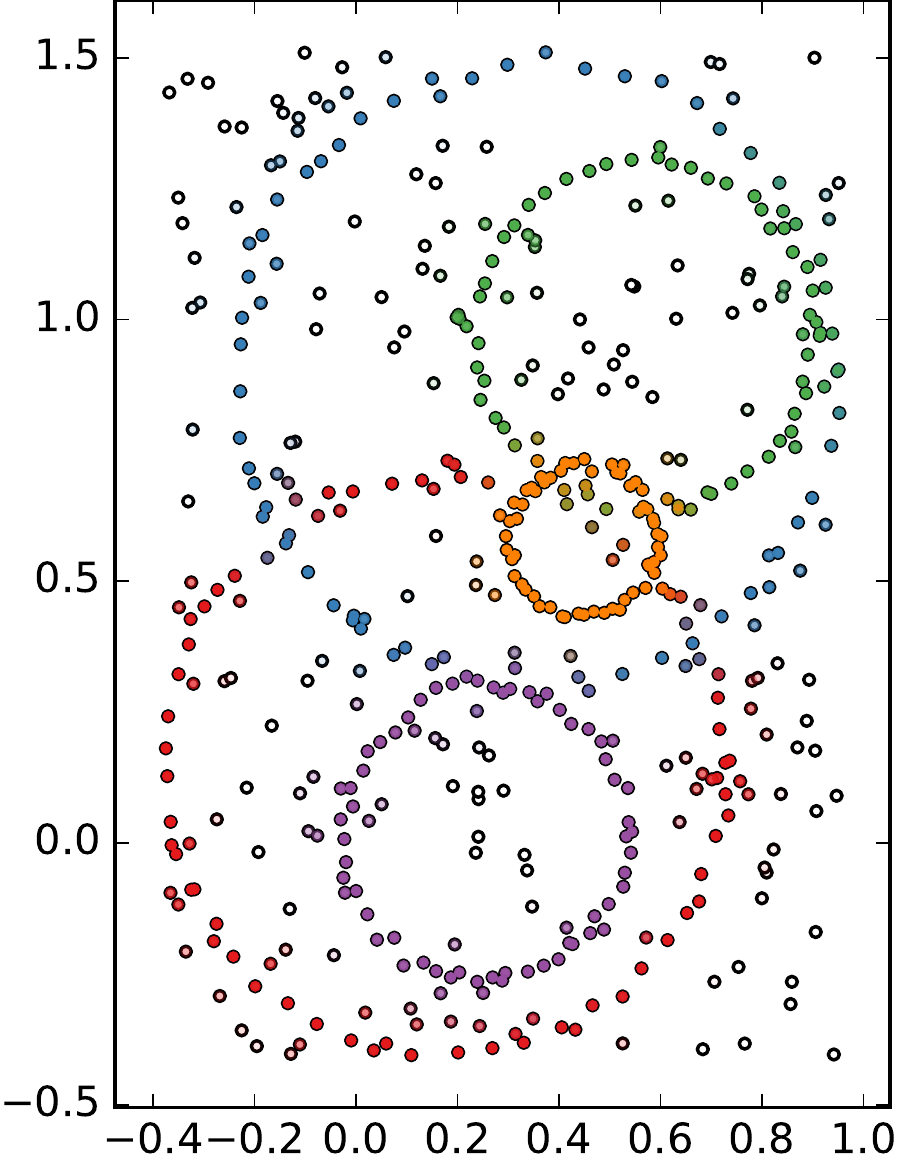}%
			\hfill%
		}

		\caption{Preference matrix (\cref{eq:preference_matrix}), before and after factorization (each NMU factors shown with a different color). We also show the computed soft-membership of points to the detected models. Points at the intersection of two circles are correctly assigned to both circles.}
		\label{fig:preference_example}
	\end{minipage}%
	\hfill%
	\begin{minipage}[b]{.64\textwidth}
		\begin{picture}(5, 5)
			\put(0,0){\includegraphics[width=\textwidth]{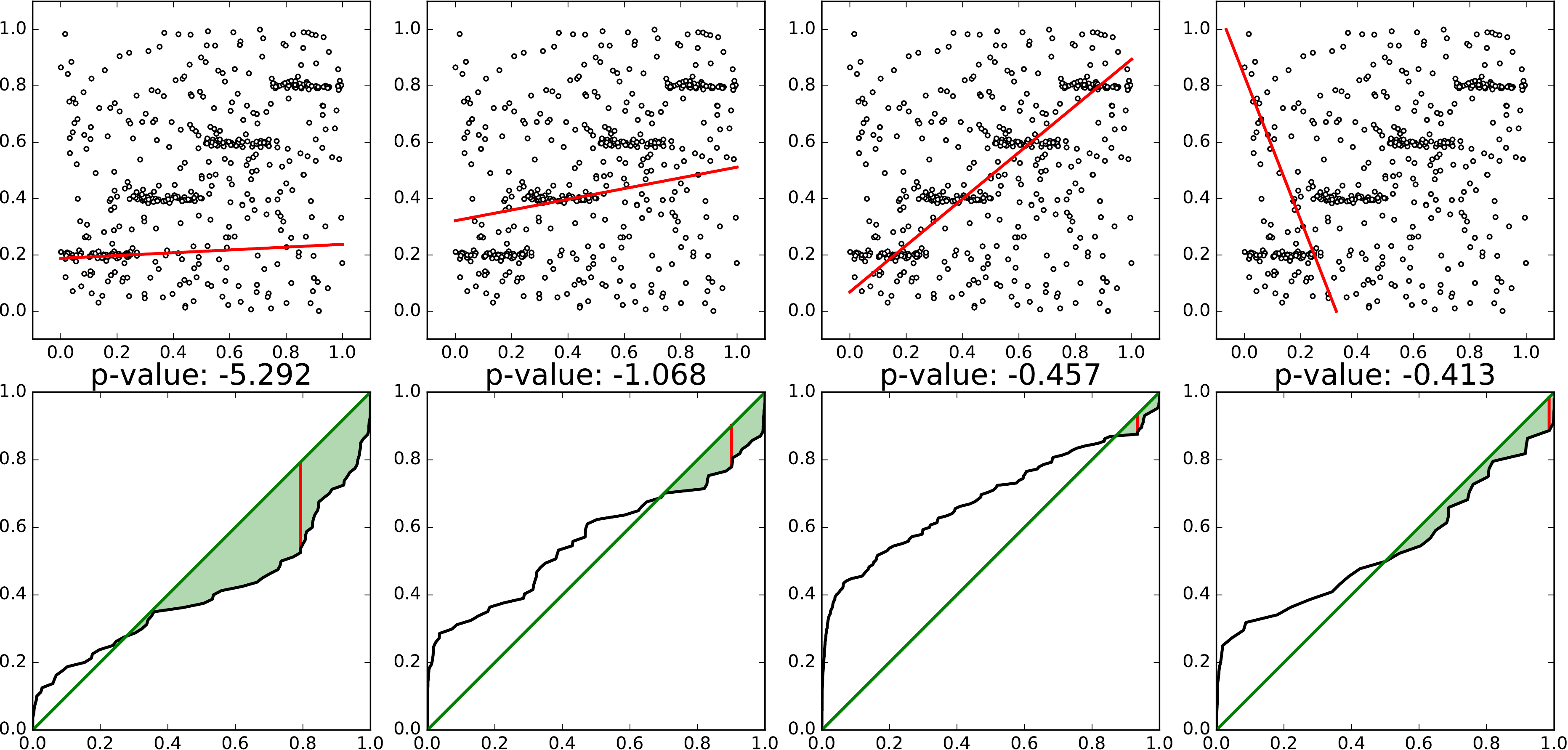}}
			\put(45,45){\begin{footnotesize}$D_-$\end{footnotesize}}
			\put(135,60){\begin{footnotesize}$D_-$\end{footnotesize}}
			\put(218,65){\begin{footnotesize}$D_-$\end{footnotesize}}
			\put(300,65){\begin{footnotesize}$D_-$\end{footnotesize}}
		\end{picture}%
		
		\caption{Examples of the statistical test presented in \cref{sec:testing} with $\sigma=0.015$ (empirical CDFs shown in the bottom row). The p-values are given in a base-10 logarithmic scale. When there is a higher concentration of points near the line than in the surrounding region (leftmost example), the p-value becomes very small. The p-value grows quickly as the line deviates from such a setting.}
		\label{fig:ks_test}
	\end{minipage}
\end{figure*}

The input of the algorithm is a collection $\set{U} = \left\{ \theta_j \right\}_{j=1}^{n}$ of model parameters, obtained with a RANSAC-like  sampling technique. There are multiple methods in the literature to obtain such a collection, e.g.,~\cite{ransac,wong11,xu1990}. In \cref{algo:random_sample} we show a basic prototypical example.
The $m \times n$ preference matrix $\mat{P}$, whose rows and columns represent respectively the $m=$ data elements $\{ \vect{x}_i \}_{i=1}^m$ and the $n=|\set{U}|$ models, with entries defined as
\begin{equation}
	(\mat{P})_{ij} = \operatorname{sm}_\mu (\vect{x}_i, \theta_j) .
	\label{eq:preference_matrix}
\end{equation}
See \cref{fig:preference_example} for a preference matrix example.

Following \cite{Tepper2014consensus,Tepper2016arse}, we pose the problem of robustly fitting multiple parametric models as a biclustering problem. In other words, we look for clusters in the space defined by Cartesian product of elements and models. In this case, the iteratively computed NMU factors are the extracted biclusters. As we will see in the following, using NMU in this context leads to an intuitive and simple interpretation of the retrieved biclusters.

\noindent\textbf{Initialization.}
In our robust fitting experiments, we found that initializing NMU-ADMM with a rank-one SVD does not provide optimal results. Perhaps unsurprisingly, the strategy chosen by RANSAC (selecting the largest consensus set) proved to be a very good initial condition to our NMU algorithm. We thus pick the largest column of $\mat{P}$, i.e.,
$j^* = \argmax_j \norm{(\mat{P})_{:j}}{1}$,
and set
\begin{subequations}
\begin{align}
	\vect{u}_0 &= (\mat{P})_{:j^*} / \norm{(\mat{P})_{:j^*}}{\infty} ,\\
	\vect{v}_0 &= \norm{(\mat{P})_{:j^*}}{\infty} \transpose{\vect{u}_0} \mat{P} / (\transpose{\vect{u}_0} \vect{u}_0) .
\end{align}
\end{subequations}

\noindent\textbf{Extracting multiple NMU factors from $\mat{P}$.}
This is a particular example of biclustering, where we know that each column of $\mat{P}$ corresponds to \emph{at most one} ground truth model. In other terms, it does not make sense to describe any given column by the combination of multiple models/biclusters. In practice, this means that once a column $(\mat{P})_{:j}$ is encoded in an NMU factor (via a positive load $(\vect{v})_j$), we can remove it entirely from $\mat{P}$.
Hence, instead of subtracting $\vect{u} \transpose{\vect{v}}$ from $\mat{P}$, we simply set the columns with positive loads to zero.

\begin{algorithm2e}[t]
	\SetKwInOut{Input}{input}
	\SetKwInOut{Output}{output}
	
	\Input{set of objects $\set{X}$, parametric function $f_\mu$.}
	\Output{pool $\set{U}$ of consensus sets.}
	
	$b \gets$ minimum number of elements necessary to uniquely characterize model $\mu$, see \cref{eq:model}\;
	
	\ForEach{$j \in \{ 1 \dots n \}$}{
		Select at random a set $\set{X}_{j}$ of $b$ elements from $\set{X}$\;
		Estimate $\theta_j$ from $\set{X}_{j}$\;
	}

	$\set{U} \gets \left\{ \theta_j \right\}_{j=1}^{n}$\;
	
	\caption{Random sampling algorithm}
	\label{algo:random_sample}
\end{algorithm2e}

\begin{figure*}[t]
	\centering
	
	\hfill%
	\begin{subfigure}{.74\textwidth}
		\begin{tabu} to \textwidth { @{\hspace{0pt}} X[r,m] *{4}{ @{\hspace{2pt}} X[r,m]} @{\hspace{0pt}} }
			\includegraphics[height=.98\linewidth]{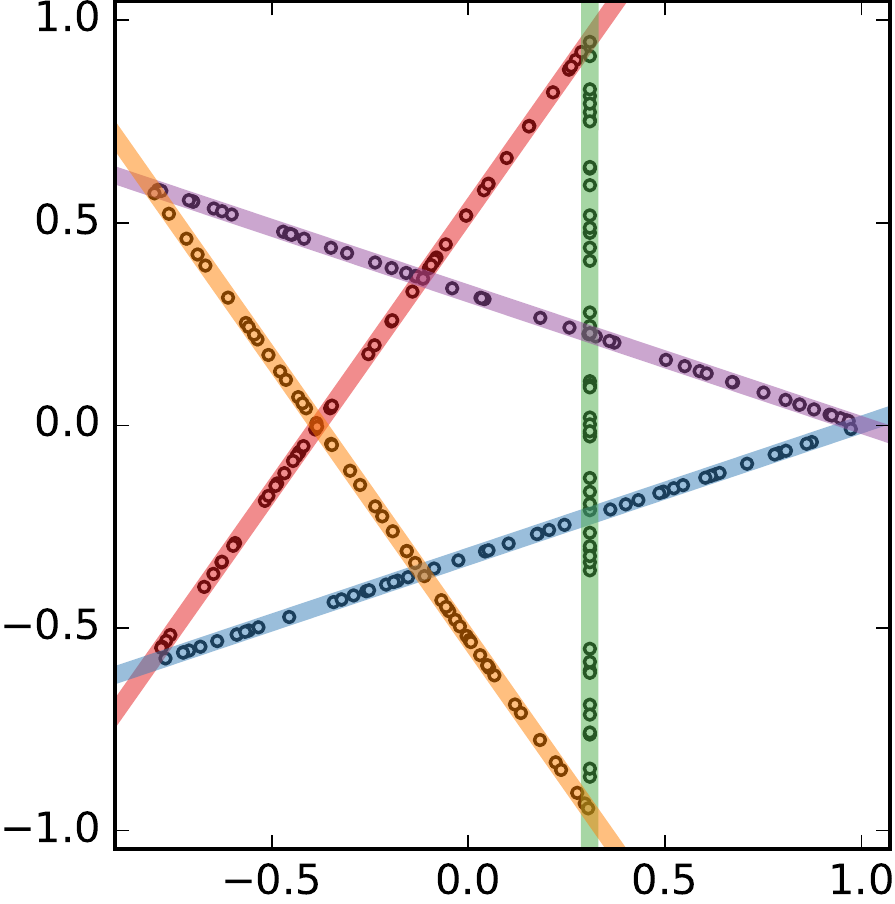}
			&%
			\includegraphics[height=.96\linewidth]{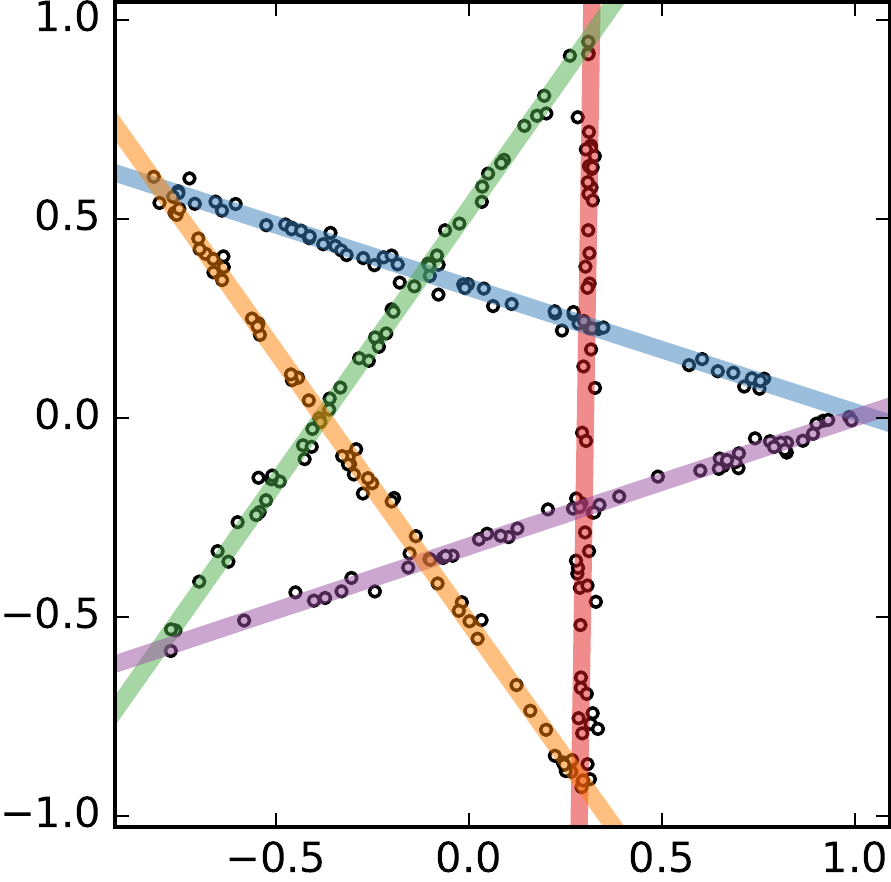}
			&%
			\includegraphics[height=.96\linewidth]{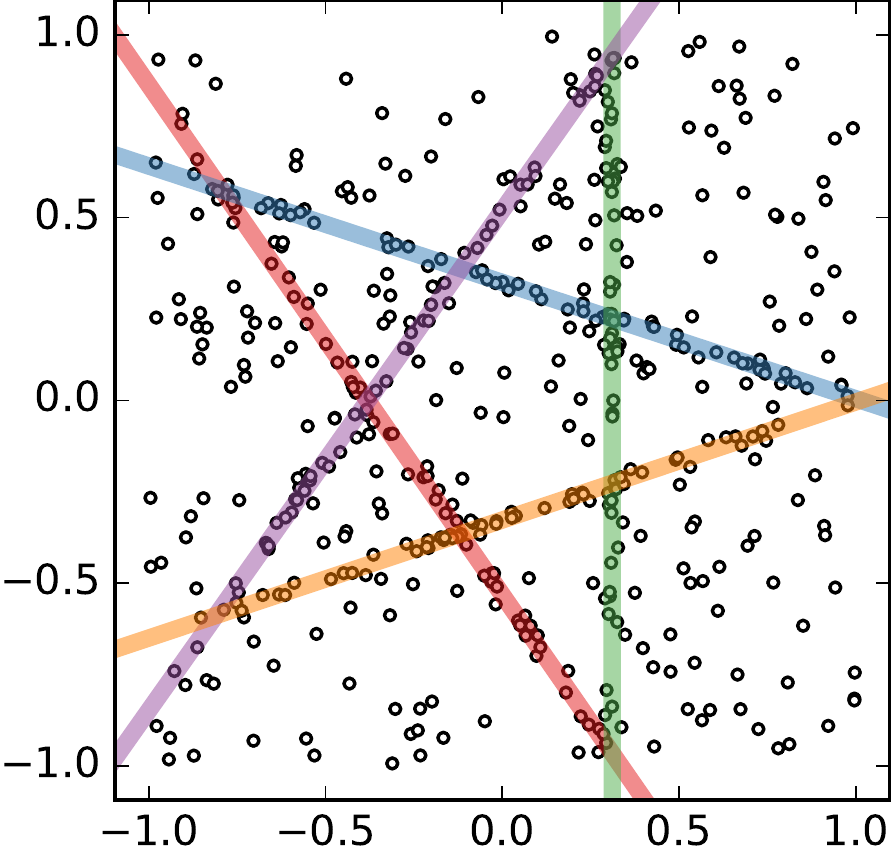}
			&%
			\includegraphics[height=.96\linewidth]{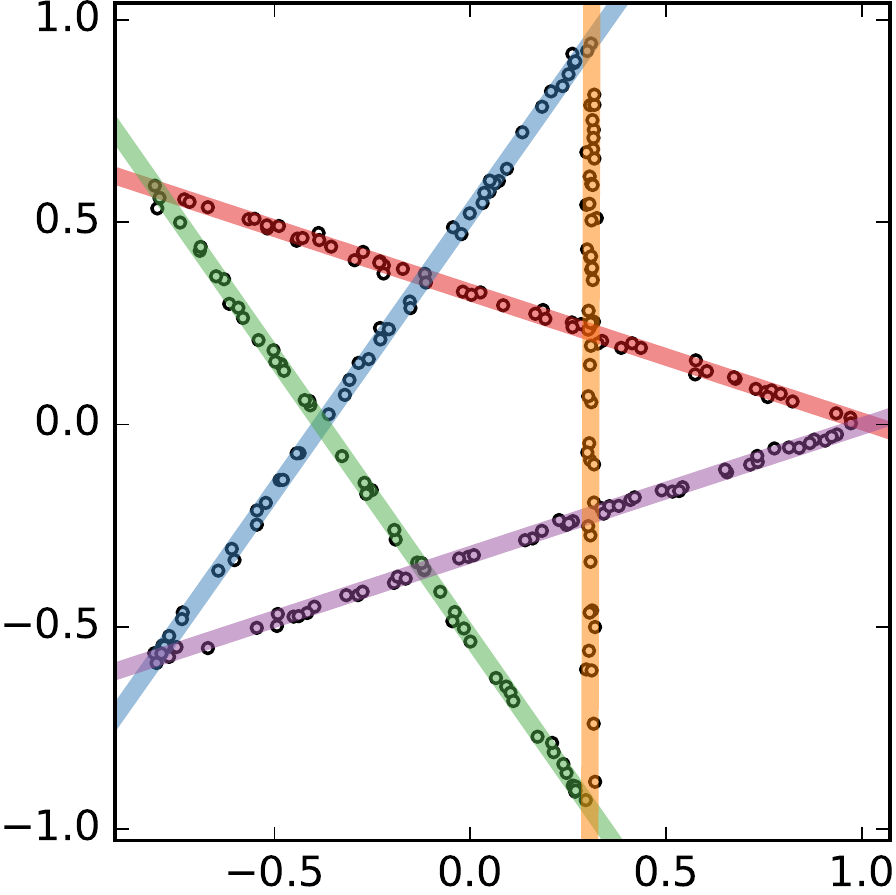}
			&%
			\includegraphics[height=.96\linewidth]{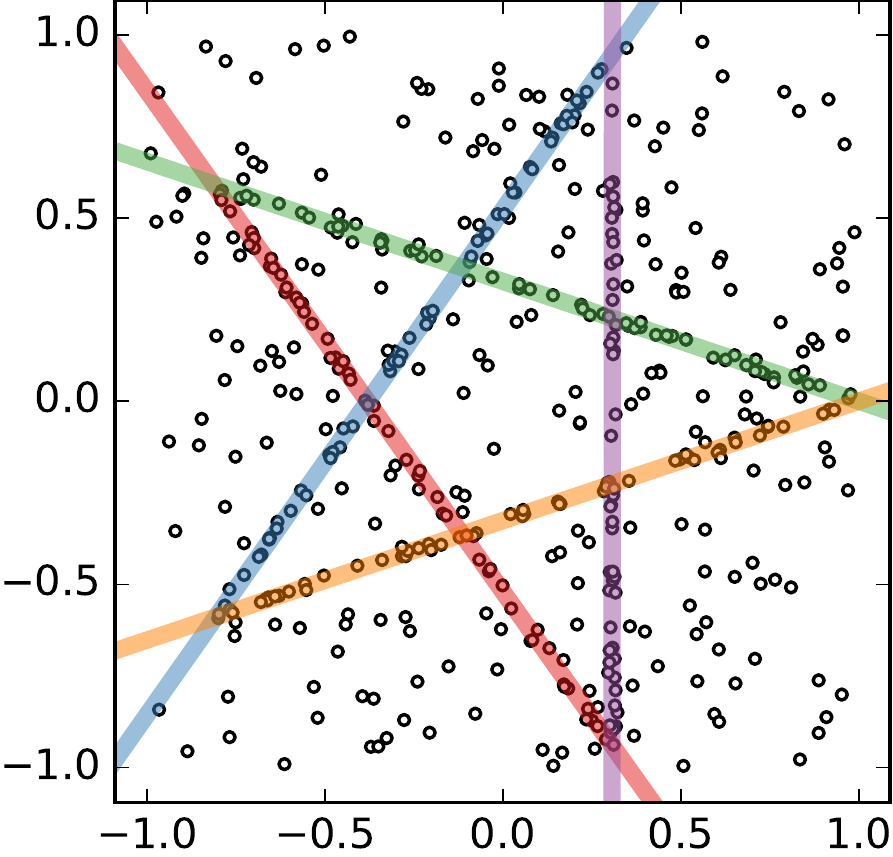}
			\\%
	
			\includegraphics[height=.96\linewidth]{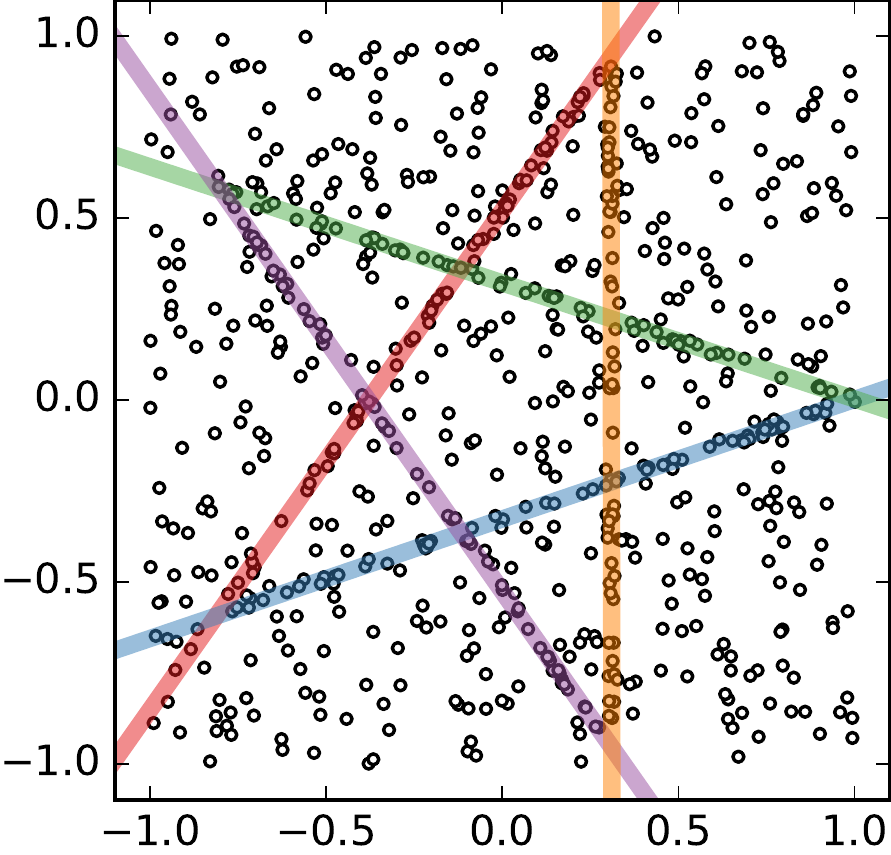}
			&%
			\includegraphics[height=.96\linewidth]{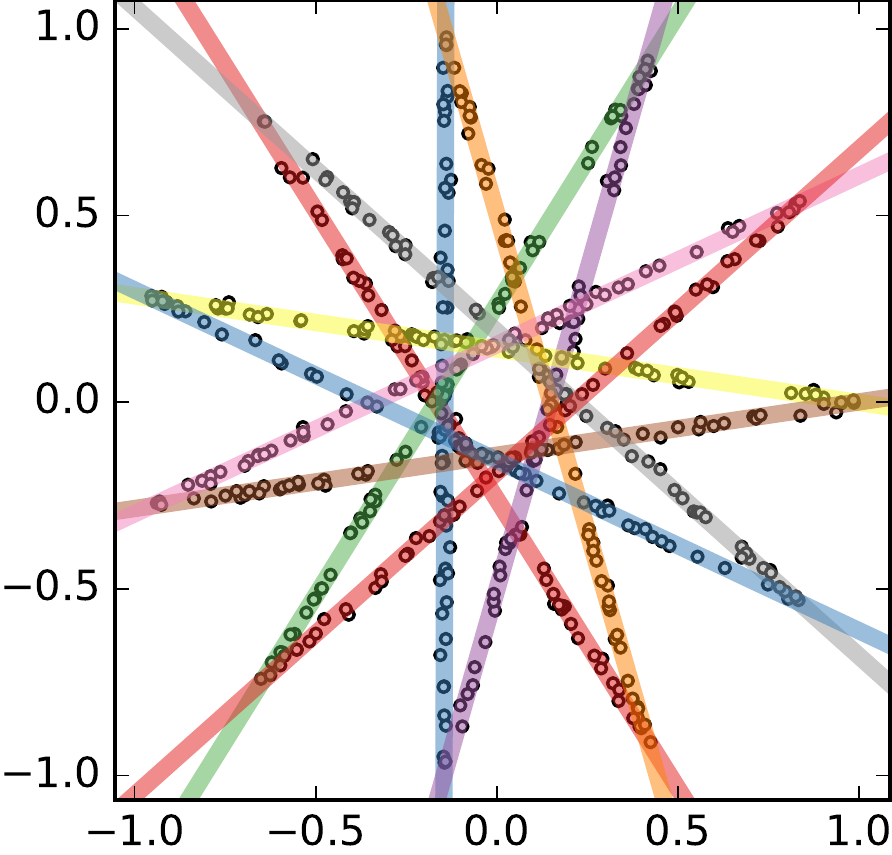}
			&%
			\includegraphics[height=.96\linewidth]{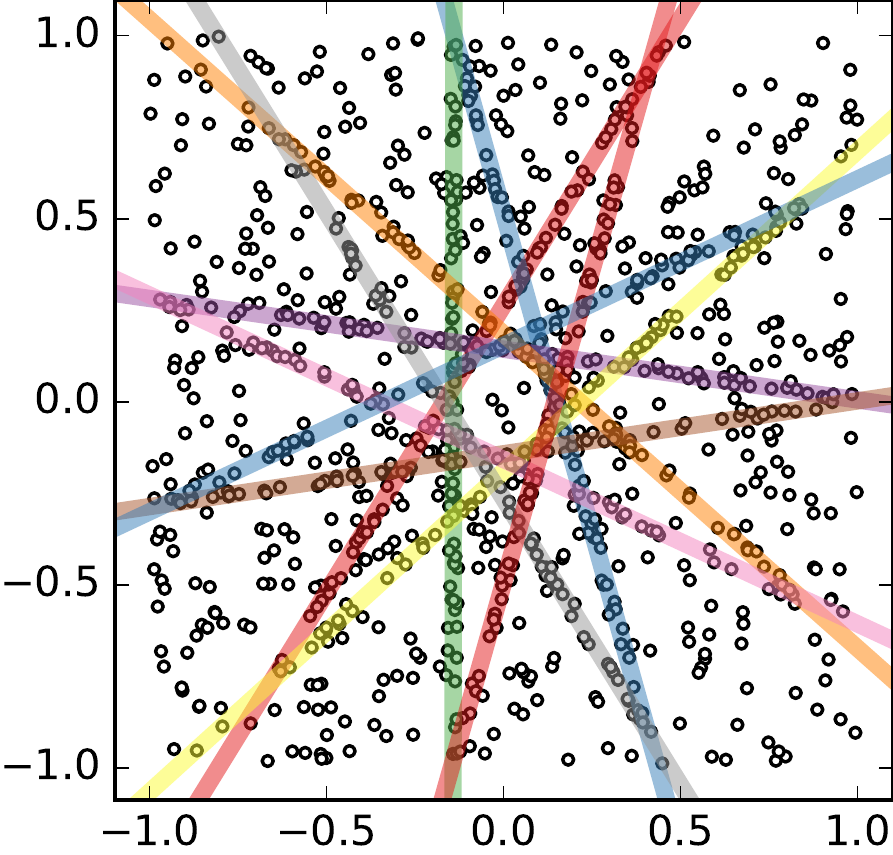}
			&%
			\includegraphics[height=.96\linewidth]{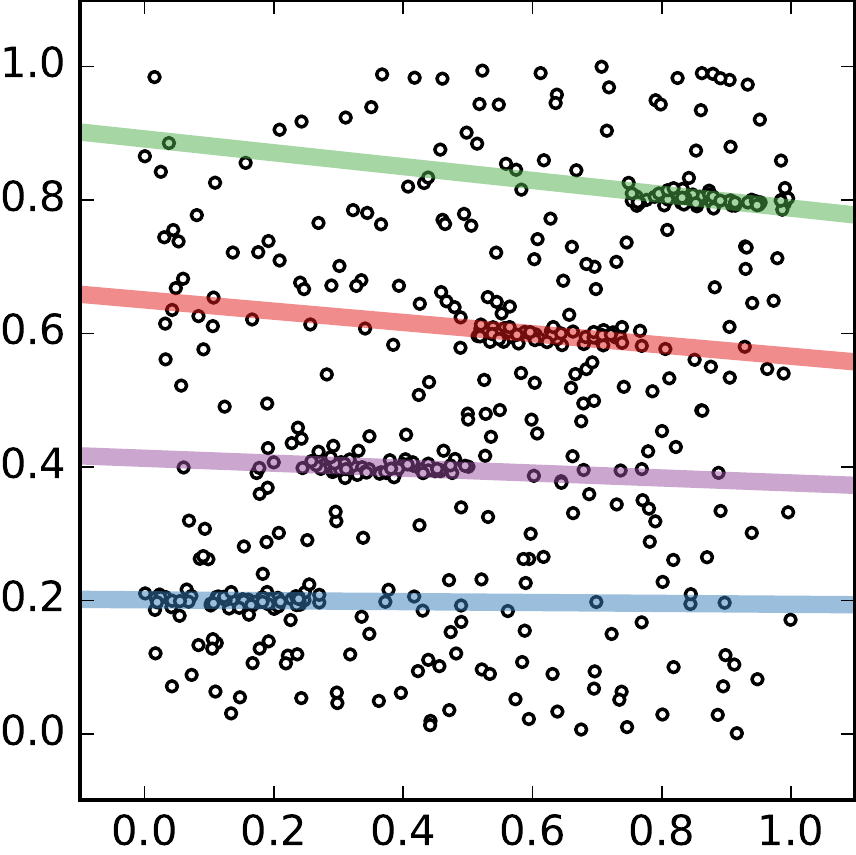}
			&%
			\includegraphics[height=.96\linewidth]{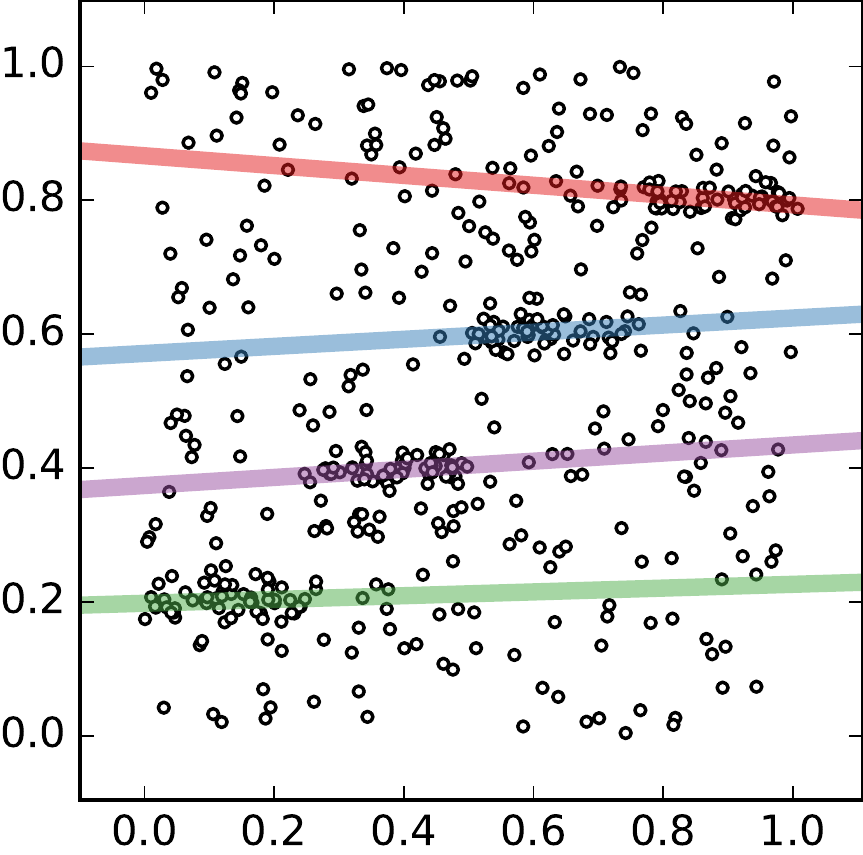}		
		\end{tabu}
	
		\caption{Line detection. In all examples $\sigma = 0.035$, except in the last one where $\sigma = 0.037$.}
		\label{fig:test_2d_lines}
	\end{subfigure}%
	\hfill
	\begin{subfigure}{.22\textwidth}
		\centering
		\includegraphics[width=.98\linewidth]{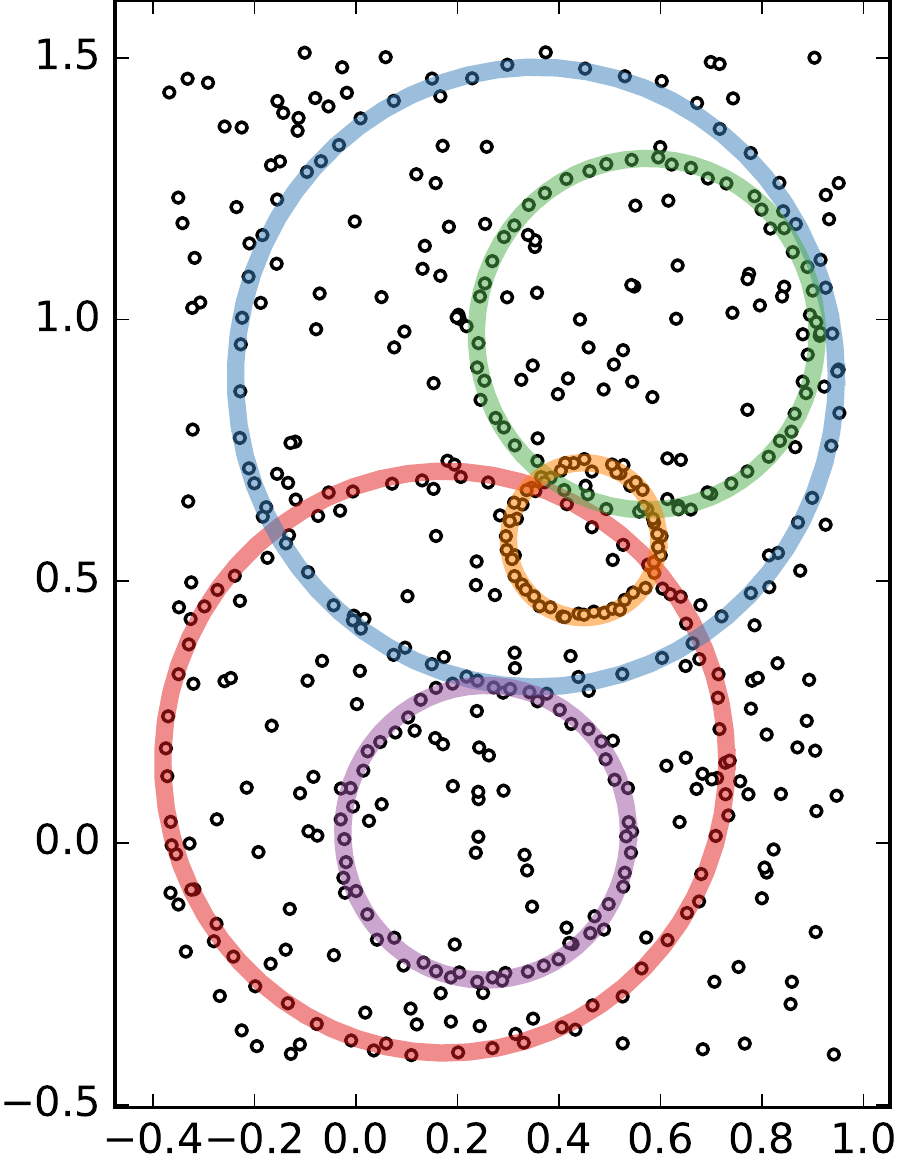}
		
		\caption{Circle detection ($\sigma = 0.047$).}
		\label{fig:test_2d_circles}
	\end{subfigure}%
	\hfill%
	
	\caption{Results on synthetic 2D datasets~\cite{toldo08}. We show with different colors the final models estimated from the extracted biclusters (some colors may repeat themselves). In every case, the models are correctly retrieved. The lines in the last example of \cref{fig:test_2d_lines} are noisier and, as one might expect, a higher $\sigma$ is needed to recover them. Estimating a circle from 3 points is much noisier than estimating a line from 2 points, as the curvature might change dramatically; hence, a higher $\sigma$ is also needed in this case.}
	\label{fig:test_2d}
\end{figure*}

\subsection{Interpretation of the NMU factors}

We consider the solution $\vect{\hat{u}}, \vect{\hat{v}}$ to \cref{eq:nmu}, provided by NMU-ADMM, with $\mat{P}$ as the input matrix.
In general terms, $(\vect{\hat{u}})_i, (\vect{\hat{v}})_j$ indicate the soft-membership of element $\vect{x}_i$ to model $\theta_j$.
If the NMU factorization considers that $(\mat{P})_{:j}$ ``agrees'' with $\vect{\hat{u}}$, then $(\vect{\hat{v}})_j > 0$; otherwise $(\vect{\hat{v}})_j = 0$.

Let us now analyze the underaproximation constraint. We first point out that since we minimize $\norm{\mat{A} - \vect{u} \transpose{\vect{v}}}{F}$, the underapproximation is as tight as possible.
Now, for any column $(\mat{P})_{:j}$, we have that $(\mat{P})_{:j} \geq \vect{\hat{u}} (\vect{\hat{v}})_j$.
Let $\vect{d}_j$ be the $m$-dimensional vector such that $(\vect{d}_j)_i = \operatorname{e}_\mu (\vect{x}_i, \theta_j)$.
Since $0 \leq (\mat{P})_{ij} \leq 1$, $0 \leq \vect{\hat{u}}, \vect{\hat{v}} \leq 1$; hence,
 \begin{subequations}
\begin{align}
	-\log\, (\mat{P})_{:j} &\leq -\log \vect{\hat{u}} - \log\, (\vect{\hat{v}})_j ,\\
	\tfrac{1}{2} (\vect{d}_j / \sigma)^2 &\leq -\log \vect{\hat{u}} - \log\, (\vect{\hat{v}})_j ,\\
	\vect{d}_j &\leq -\sigma \log^{1/2} \vect{\hat{u}}^2 -\sigma \log^{1/2}\, (\vect{\hat{v}})_j^2
\end{align}
\end{subequations}
where the logarithm is taken element-wise. The nonnegative scalar $-\sigma \log^{1/2}\, (\vect{\hat{v}})_j^2$ provides a global safety margin over the distances $\vect{d}_j$ to ensure that they are included in the solution. The higher the agreement, the lower the margin (if $(\vect{\hat{v}})_j = 0$, we have an infinite margin). A similar pattern is observed for $- \sigma \log^{1/2} \vect{\hat{u}}^2$, in this case individualizing the margin for each data point $\vect{x}_i$ (whether more or less leniency is needed to include it in the solution).

This interpretation of the values of $\vect{\hat{u}}$ and $\vect{\hat{v}}$, being derived from the underapproximation constraint, is not present in other formulations of the robust fitting problem~\cite{denitto2016,Tepper2014consensus}.

\subsection{Computing model instances from NMU factors}

Once the NMU factors $\vect{\hat{u}}_t, \vect{\hat{v}}_t$ are found, we estimate the corresponding $\hat{\theta}_t$ by weighted least-squares regression, i.e.,
\begin{equation}
\hat{\theta}_t = \argmin_{\theta}  \sum_{i=1}^{m} (\vect{\hat{u}}_t)_i  \big[ \operatorname{e}_\mu (\vect{x}_i, \theta) \big]^2 .
\label{eq:leastSquares}
\end{equation}
Each data point contributes to the estimation of the final model instance proportionally to its soft-membership to the factor $\vect{\hat{u}}$.

\subsection{Filtering candidates with statistical testing}
\label{sec:testing}

Not all of the extracted NMU factors will correspond to meaningful models; in particular, the very last factors will correspond to noise in $\mat{P}$ (i.e., spurious models).
We use statistical testing to determine whether each NMU factor $\vect{\hat{u}}_t, \vect{\hat{v}}_t$ is to be kept in the final result.

Let $F_t$ be the empirical cumulative distribution function (CDF) of the values $\operatorname{sm}_\mu (\vect{x}_i, \hat{\theta}_t)$ and $F_U$ denote the CDF for the uniform distribution over the interval $[0, 1]$. Following Kuiper's statistical test, we define the statistic
\begin{equation}
	D_{-} = \max_{0 \leq x \leq 1} F_U(x) - F_t(x) .
\end{equation}
Since Kuiper's test is closely related to the Kolgomorov-Smirnov test, the statistic $D_{-}$ follows a Kolmogorov-Smirnov CDF. In \cref{fig:ks_test}, we can see several examples of its value for different lines. The p-value of $D_{-}$ is a good indicator of a high concentration of values near one. Hence, it is a good indicator of a high concentration of elements near the evaluated instance model.

The null hypothesis is rejected if $D_{-} > K_\alpha$ where $K_\alpha$ is found from $\Pr (K > K_\alpha) = \alpha.$
The value $\alpha$ is automatically set to $1 / \binom{m}{b}$ where $\binom{\cdot}{\cdot}$ is the combinatorial number and $b$ is the minimum number of elements necessary to estimate/instantiate a model~\cite{desolneux08}.

This test can also be used as a preprocessing step.
The standard random sampling approach to multiple model estimation generates many good model instances (composed of inliers), but also generates many bad models (composed mostly of outliers).
Then, prior to computing the NMU factors, we discard the columns of $(\mat{P})_{:j}$ for which the null hypothesis is not rejected.
Here, we also set $\alpha=1 / \binom{m}{b}$.

\subsection{From NMU factors to independent models}
\label{sec:redundancy}

In our framework, models are allowed to share elements as there are no orthogonality constraints between successive NMU factors. However, if two distinct factors $\vect{\hat{u}}_t$ and $\vect{\hat{u}}_{t'}$ are too similar, the estimated models $\hat{\theta}_t$ and $\hat{\theta}_{t'}$ will be too similar, producing redundant output information.

To avoid this, we compute the correlation $c_{t, t'}$ between every pair of factors $t, t'$.
We consider two factors independent from each other if this quantity is lower than $0.6$ (i.e., an angle between them is above~53\textdegree approx., which seems to a reasonable universal choice).
We build a graph by linking together the vertices $t$ and $t'$ if $c_{t, t'} > 0.6$.
Notice that the number of factors is a very small number here, making the graph small. We then list all maximal independent sets (MIS) in this graph. Each MIS is a set of mutually compatible factors/models.

We seek to determine which MIS of models/factors is the one better describing the data. Many different techniques can be used for this task, such as the Bayesian and Akaike information criteria or minimum description length. We follow a simplified approach: for each MIS, we consider the geometric mean of the p-values associated with its models (computed with the previously described statistical test); we return the MIS with minimum geometric mean. We note that results could be improved by using a more sophisticated selection method.

Optionally and if required by the application, as the last step of our algorithmic pipeline, we can force the models to have an empty intersection. There are many alternative ways to address this assignment problem. In this work, we simply assign elements in the intersection of several models to the closest model in distance (see \cref{eq:pointModelDistance}).

\section{Experimental results}
\label{sec:results}

\begin{figure}[t]
	\centering

	\begin{footnotesize}
	\begin{tabu} to \columnwidth {@{\hspace{0pt}} X[c,m] @{\hspace{0pt}} X[c,m] @{\hspace{4pt}} X[19,c,m] @{\hspace{0pt}}}
		\begin{sideways}Fundamental matrices\end{sideways} &
		\begin{sideways}`biscuitbookbox'\end{sideways} &
		\includegraphics[width=\linewidth]{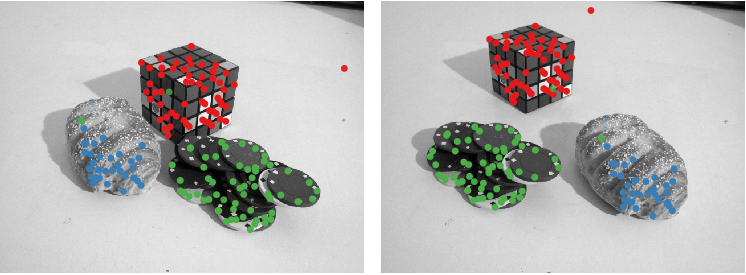}\\
		\\[-4pt]
		\begin{sideways}Homographies\end{sideways} &
		\begin{sideways}`oldclassicswing'\end{sideways} &
		\includegraphics[width=\linewidth]{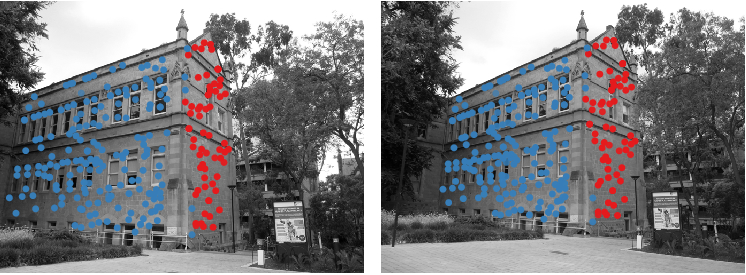}%
	\end{tabu}
	\end{footnotesize}
	
	\caption{Two prototypical examples of the results in \cref{tab:adelaide} with misclassification errors $3.09\%$ (top) and $1.58\%$ (bottom).}
	\label{fig:adelaide_nice}
\end{figure}

\begin{figure}[t]
	\centering
	
	\begin{footnotesize}
		\includegraphics[width=.9\columnwidth]{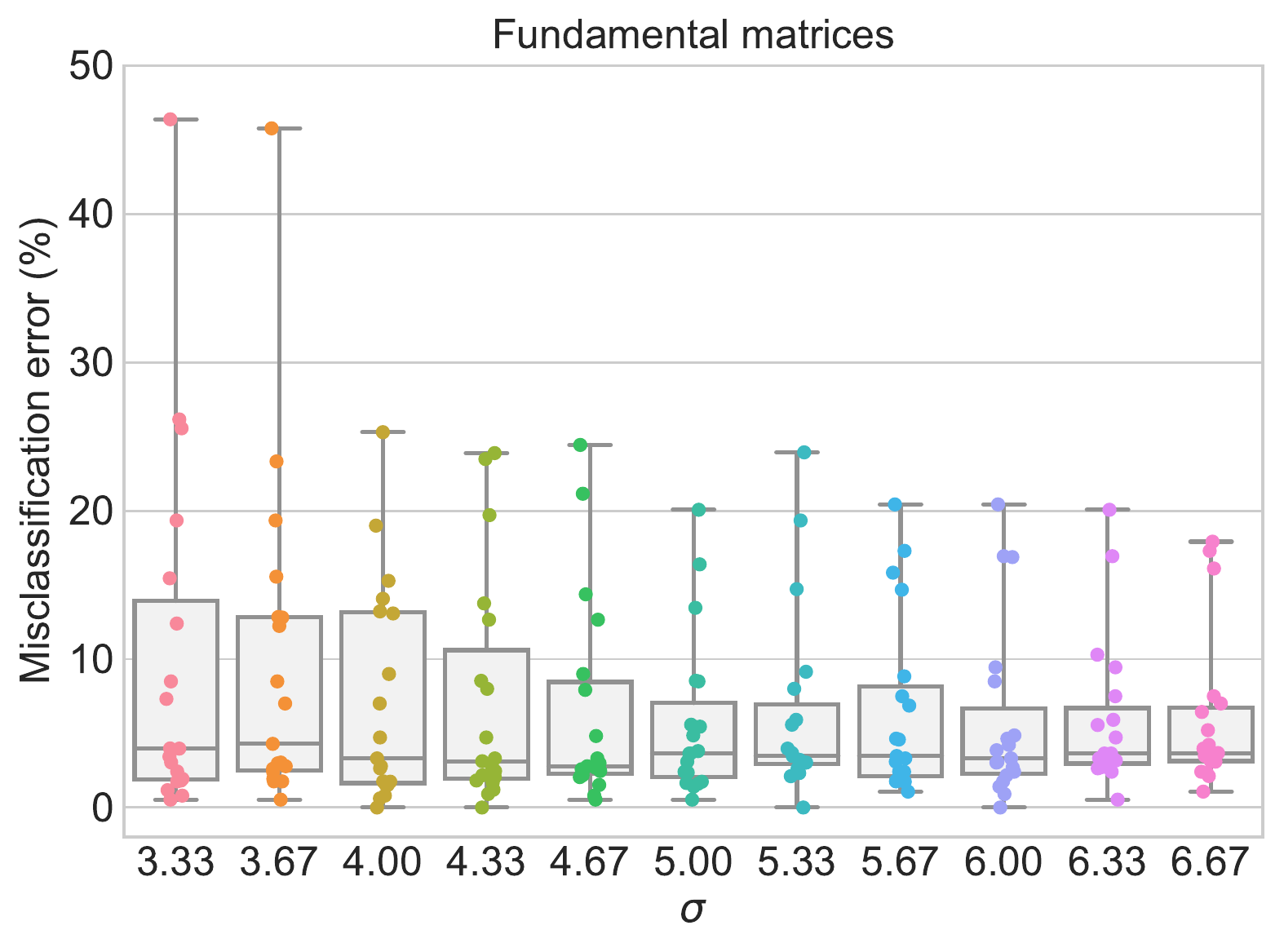}\\
		\includegraphics[width=.9\columnwidth]{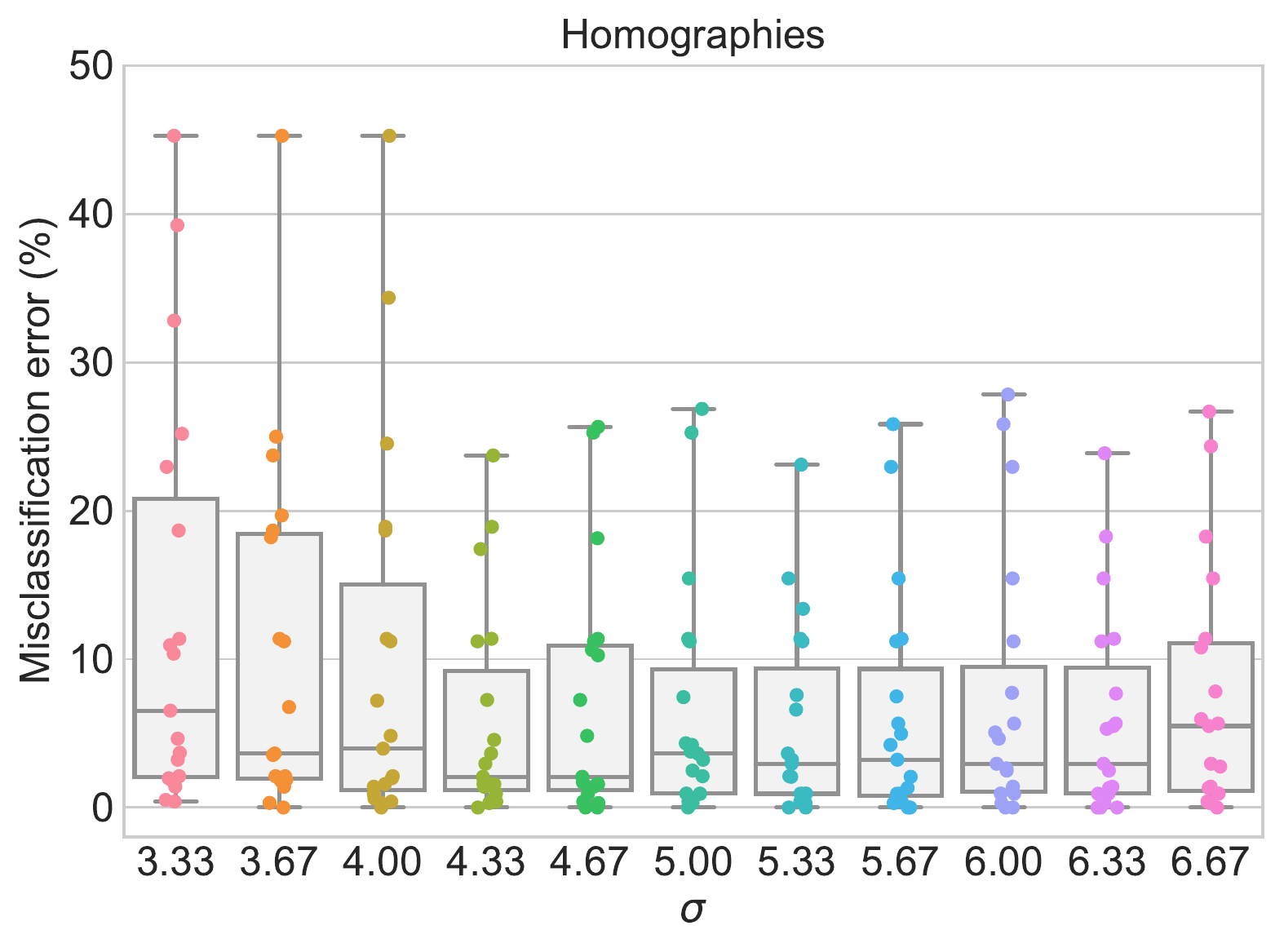}%
	\end{footnotesize}
	
	\caption{Results on the AdelaideRMF dataset~\cite{wong11} for different $\sigma$ (\cref{eq:soft_membership}). A small horizontal jitter was added to each colored point in the plot to improve visualization. The results are roughly stable to the choice of $\sigma$, making this selection less critical than in the hard-membership approaches, i.e., with a binary preference matrix, e.g.~\cite{Tepper2016arse,toldo08}.}
	\label{fig:adelaide_sigma}
\end{figure}

We start our experimental evaluation with a few small synthetic datasets \cite{toldo08} where 2D points are arranged forming lines and circles. The results are shown in \cref{fig:test_2d}, where all models are clearly detected. 
In this case, we use \cref{algo:random_sample} as a sampling strategy, all points being treated as equiprobable (we sample from a uniform distribution).

The proposed approach can correctly recover overlapping models, see \cref{fig:test_2d}. This can be observed in detail by comparing the results in \cref{fig:test_2d_circles} with its preference matrix in \cref{fig:preference_example}. This is an intrinsic limitation of previous state-of-the-art competitors such as J-linkage~\cite{toldo08}, T-linkage~\cite{magri2014}, RPA~\cite{magri2015} and most parametric fitting techniques, which are generally based on partitioning (clustering) the dataset.

We next estimate multiple fundamental matrices (moving camera and moving objects) and multiple homographies (moving planar objects) on the images in the AdelaideRMF dataset~\cite{wong11}. This is a standard dataset in the literature (e.g.,~\cite{denitto2016, magri2014,magri2015,Pham2014,Tepper2016arse,wong11}), but as pointed out in~\cite{Tepper2016arse} it contains a non-negligible quantity of errors in its ground truth. This implies that, beyond some point, an improvement in the actual performance might not necessarily reflect itself as an improvement compared to the ground truth.

\begin{figure*}[t]
	\centering

	\begin{footnotesize}
	\begin{tabu} to \textwidth { @{\hspace{0pt}} X[c,m] X[15,c,m] *{2}{| X[16,c,m]} @{\hspace{0pt}} }
		\begin{sideways}
			\shortstack{
				Fund. mat.\\
				`boardgame'
			}
		\end{sideways} &
		\includegraphics[width=\linewidth]{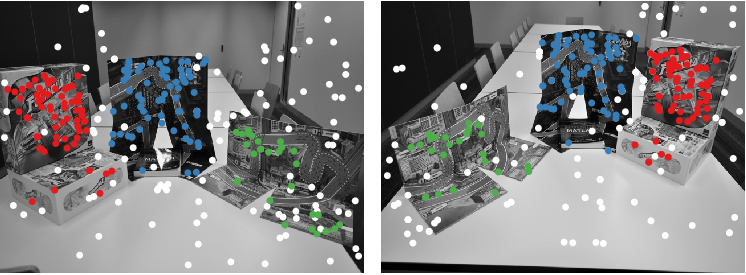} &
		\includegraphics[width=\linewidth]{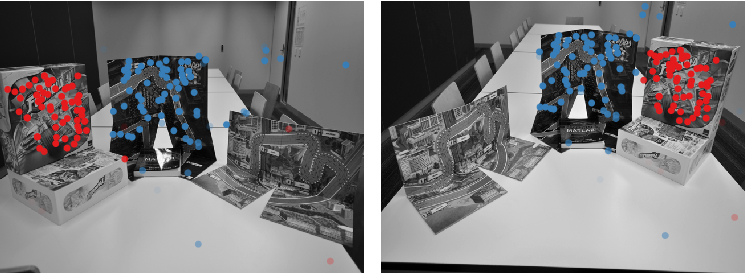} &
		\includegraphics[width=\linewidth]{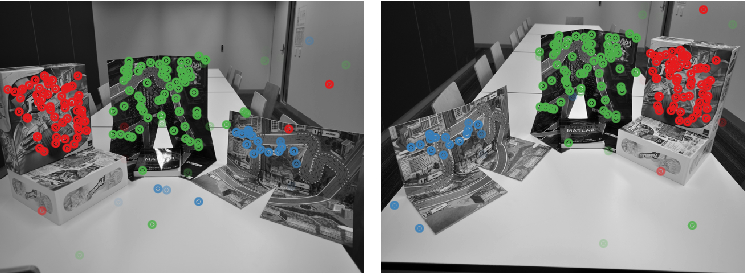}\\
		
		\\[-4pt]
	
		\begin{sideways}
			\shortstack{
				Homographies\\
				`johnsonb'
			}
		\end{sideways} &	
		\includegraphics[width=\linewidth]{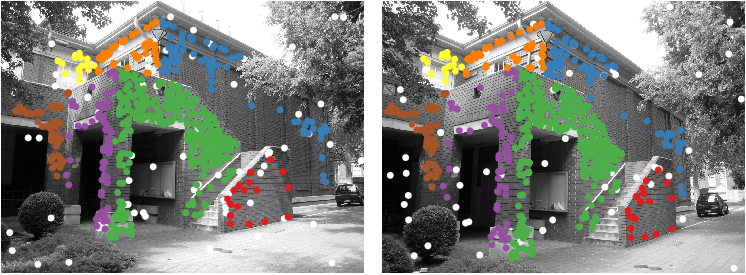} &		
		\includegraphics[width=\linewidth]{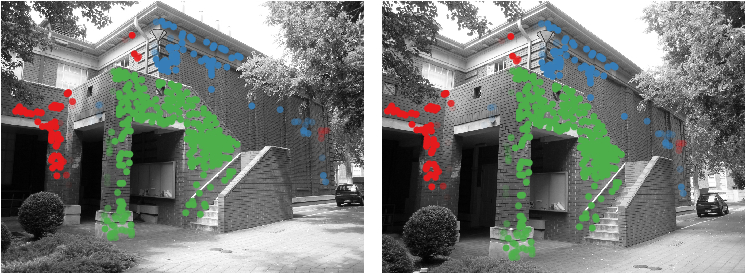} &
		\includegraphics[width=\linewidth]{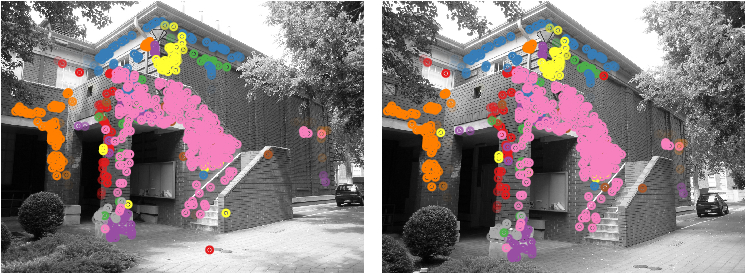} \\
	\end{tabu}
	\end{footnotesize}
	
	\caption{Two examples in which the method exhibits a higher misclassification error (ME) in \cref{tab:adelaide}. From left to right: ground truth hard assignments, RS-NMU soft assignments, and RS-NMU soft assignments with no statistical testing performed (in pre- nor in post-processind).
	Top row: the ME goes from $20.07\%$ (center) to $12.54\%$ (right); the main diference is the appearence of an additional model that had been discarded for not passing the statistical test.
	Bottom row: the ME goes from $23.73\%$ (center) to $32.82\%$ (right); a few `desired' groups appear (e.g., red dots), but accompanied with some `undesired' groups (e.g., yellow dots). The problem of differentiating the `desired' from the `undesired' groups is a very hard when they contain few points.}
	\label{fig:adelaide_bad}
\end{figure*}

As standard in the literature~\cite{denitto2016,magri2014,magri2015,Pham2014,Tepper2016arse} we use the misclassification error (ratio of misclassified points) to measure performance.
While the ground truth assigns elements to groups in a hard fashion, the values $\operatorname{sm}_\mu (\vect{x}_i, \hat{\theta}_t)$ are continuous in $[0, 1]$. To compute this error, we simply binarize them by considering that an element $\vect{x}_i$ belongs to group $t$ if $\operatorname{sm}_\mu (\vect{x}_i, \hat{\theta}_t) > 0$.

To show that the proposed method is agnostic to the random sampling technique, we use MultiGS~\cite{chin2012} for these experiments.\footnote{Interestingly, a recent technique~\cite{BenArtzi2016} has reduced to two the minimal sample size (see \cref{algo:random_sample}) for fundamental matrices; its use would significantly reduce the computational complexity of the sampling step, imposing an upper bound of $O(m^2)$ to the total number of possible combinations.} Of the total running time, MultiGS uses on average $98\%$, while the remaining $2\%$ corresponds to the proposed ADMM algorithm for NMU.

The results on the full dataset are presented in \cref{tab:adelaide}.
For this experiment, we term the approach proposed in \cref{sec:mpme} as RS-NMU. We clarify that RPA~\cite{magri2015} and FABIA~\cite{denitto2016} achieve good results; however, to yield these performances, they use the \emph{ground truth} number of models as an input. The remaining methods, as well as the method here proposed, automatically estimate this number.

To estimate fundamental matrices, see \cref{tab:adelaide_fundamental}, RS-NMU outperforms all its competitors, even being slightly better in median than RPA. To estimate homographies, see \cref{tab:adelaide_homography}, the differences with the other methods are even higher, with RS-NMU clearly outperforming all of them. These observations hold true, even if the $\sigma$ is not accurately set (compare the last two columns in \cref{tab:adelaide_fundamental,tab:adelaide_homography}).
In \cref{fig:adelaide_nice} we show two examples of the RS-NMU results.
This improved performance comes from a lightweight algorithm, using approx.~$2\%$ of the total running time.

As one might expect, there is not a universal value of $\sigma$ that works best for all the image pairs, see for example `breadcube' and `biscuit' in \cref{tab:adelaide_fundamental}.

\begin{table*}[p]		
	\newcolumntype{Z}{S[table-format=2.2,round-integer-to-decimal=true,round-mode=places]}
	\tabucolumn{Z}
	
	\begin{subtable}{\textwidth}
		\centering
		\begin{tabu}{l *{8}{Z}}
			\hline
			
			& {RSE} & {ARSE} & {T-linkage} & {RCMSA} & {RPA} & {FABIA} & {RS-NMU} & {RS-NMU} \\
			\\[-16pt]
			& {\cite{Tepper2016arse}} & {\cite{Tepper2016arse}} & {\cite{magri2014}} & {\cite{Pham2014}} & {\cite{magri2015}} & {\cite{denitto2016}} & {$\sigma=5.00$} & {$\sigma=6.00$} \\

			\hline			

			biscuit           & 22.60 & \firstcolor 0.90  & 16.93 & 14.00 & 1.15  & 0.00  & 5.45  & \secondcolor 0.91  \\
			biscuitbook       & 3.50  & \firstcolor 1.20  & 3.23  & 8.41  & 3.23  & 1.32  & 2.35  & \secondcolor 1.76  \\
			biscuitbookbox    & 3.90  & \firstcolor 1.90  & 3.10  & 16.92 & 3.88  & 3.86  & 3.09  & \secondcolor 2.70  \\
			boardgame         & \secondcolor 15.40 & \firstcolor 15.00 & 21.43 & 19.80 & 11.65 & 8.96  & 20.07 & 20.43 \\
			book              & 3.80  & 2.20  & 3.24  & 4.32  & 2.88  & 0.00  & \secondcolor 0.53  & \firstcolor 0.00  \\
			breadcartoychips  & 27.70 & 28.60 & \secondcolor 14.29 & 25.69 & 7.50  & 4.22  & \firstcolor 3.80  & 16.88 \\
			breadcube         & 2.60  & \firstcolor 0.90  & 19.31 & 9.87  & 4.58  & 19.42 & \secondcolor 1.65  & 3.31  \\
			breadcubechips    & 15.70 & 14.80 & 3.48  & 8.12  & 5.07  & 0.87  & \firstcolor 1.74  & \secondcolor 2.17  \\
			breadtoy          & \firstcolor 1.40  & \secondcolor 2.20  & 5.40  & 3.96  & 2.76  & 19.62 & 4.86  & 4.86  \\
			breadtoycar       & 25.00 & 25.00 & 9.15  & 18.29 & 7.52  & 0.60  & \firstcolor 2.41  & \secondcolor 4.22  \\
			carchipscube      & 12.80 & 12.80 & 4.27  & 18.90 & 6.50  & 1.52  & \secondcolor 3.64  & \firstcolor 3.03  \\
			cube              & \firstcolor 2.40  & 12.50 & 7.80  & 8.14  & 3.28  & 1.66  & \secondcolor 2.64  & 4.64  \\
			cubebreadtoychips & 13.70 & 14.30 & \secondcolor 9.24  & 13.27 & 4.99  & 1.07  & 13.46 & \firstcolor 3.06  \\
			cubechips         & 2.90  & 3.60  & 6.14  & 7.70  & 4.57  & 0.53  & \firstcolor 1.41  & \firstcolor 1.41  \\
			cubetoy           & 3.80  & 4.60  & 3.77  & 5.86  & 4.04  & 2.21  & \firstcolor 1.61  & \secondcolor 2.41  \\
			dinobooks         & 23.60 & 18.30 & 20.94 & 23.50 & 15.14 & 9.72  & \firstcolor 16.39 & \secondcolor 16.94 \\
			game              & \firstcolor 0.40  & \secondcolor 0.90  & 1.30  & 5.07  & 3.62  & 0.00  & 5.58  & 3.86  \\
			gamebiscuit       & 9.90  & \firstcolor 4.30  & 9.26  & 9.37  & 2.57  & 2.44  & \secondcolor 8.54  & 9.45  \\
			toycubecar        & \firstcolor 8.10  & \firstcolor 8.10  & 15.66 & 13.81 & 9.43  & 9.50  & 8.50  & 8.50  \\
			
			\hline
			
			Mean              & 10.48 & 9.06  & 9.37  & 12.37 & 5.49  & 4.61  & \firstcolor 5.72  & \secondcolor 5.82  \\
			STD               & 8.97  & 8.57  & 6.66  & 6.58  & 3.48  & 6.13  & 5.44  & 5.96  \\
			Median            & 8.10  & 4.60  & 7.80  & 9.87  & 4.57  & 1.66  & \secondcolor 3.64  & \firstcolor 3.31 \\

			\hline
		\end{tabu}
		
		\caption{Misclassification error ($\%$) for the estimation of multiple fundamental matrices.}
		\label{tab:adelaide_fundamental}
	\end{subtable}
	\\[4pt]
	\begin{subtable}{\textwidth}			
		\centering
		\begin{tabu}{l *{8}{Z}}
			\hline
			
			& {RSE} & {ARSE} & {T-linkage} & {RCMSA} & {RPA} & {FABIA} & {RS-NMU} & {RS-NMU} \\
			\\[-16pt]
			& {\cite{Tepper2016arse}} & {\cite{Tepper2016arse}} & {\cite{magri2014}} & {\cite{Pham2014}} & {\cite{magri2015}} & {\cite{denitto2016}} & {$\sigma=4.33$} & {$\sigma=5.33$} \\
			
			\hline

			barrsmith       & \firstcolor 3.40  & \secondcolor 4.70      & 49.79 & 20.14 & 36.31 & 29.88             & 11.20             & 11.20 \\
			bonhall         & 29.20 & 38.00     & 21.84 & 19.69 & 41.67 & 24.02             & \secondcolor 17.42             & \firstcolor 13.39 \\
			bonython        & 1.60  & 5.20      & 11.92 & 17.79 & 15.89 & 6.82              & \firstcolor 0.00              & \firstcolor 0.00  \\
			elderhalla      & 6.10  & 3.70      & 10.75 & 29.28 & 0.93  & 3.04              & \firstcolor 0.93              & \firstcolor 0.93  \\
			elderhallb      & 12.20 & 20.40     & 31.02 & 35.78 & 17.82 & 18.63             & \firstcolor 11.37             & \firstcolor 11.37 \\
			hartley         & 1.60  & 1.30      & 21.90 & 37.78 & 17.78 & 23.59             & \secondcolor 0.94              & \firstcolor 0.62  \\
			johnsona        & 13.90 & 16.40     & 34.28 & 36.73 & 10.76 & 17.96             & \secondcolor 4.56              & \firstcolor 3.22  \\
			johnsonb        & \firstcolor 15.10 & 17.60     & 24.04 & \secondcolor 16.46 & 26.76 & 24.50             & 23.27             & 23.11 \\
			ladysymon       & 3.50  & \firstcolor 2.60      & 24.67 & 39.50 & 24.67 & 11.81             & \secondcolor 2.95              & \secondcolor 2.95  \\
			library         & 2.40  & \firstcolor 0.90      & 24.53 & 40.72 & 31.29 & 20.47             & 1.40              & \secondcolor 0.93  \\
			napiera         & 8.20  & 11.00     & 28.08 & 31.16 & 9.25  & 21.85             & \firstcolor 2.98              & \secondcolor 5.30  \\
			napierb         & 18.10 & 18.60     & \firstcolor 13.50 & 29.40 & 31.22 & 36.68             & 18.92             & \secondcolor 15.83 \\
			neem            & 4.30  & 7.80      & 25.65 & 41.45 & 19.86 & 11.20             & \firstcolor 2.07              & \secondcolor 2.90  \\
			nese            & 1.20  & 2.10      & 7.05  & 46.34 & 0.83  & 4.92              & \firstcolor 0.00              & \secondcolor 0.39  \\
			oldclassicswing & \secondcolor 1.70  & 2.50      & 20.66 & 21.30 & 25.25 & 7.92              & \firstcolor 1.58              & 2.11  \\
			physics         & 28.20 & 22.30     & 29.13 & 48.87 &  0.00  &  0.00              & \firstcolor 1.89              & \firstcolor 2.83  \\
			sene            & \firstcolor 0.40  & 0.80      & 7.63  & 20.20 & 0.42  & 2.20              & 1.60              & \firstcolor 0.40  \\
			unihouse        & 7.60  & 11.20     & 33.13 & \firstcolor 2.56  &  5.21  & 15.76             & \secondcolor 7.25              & 7.58  \\
			unionhouse      & 0.90  & 0.90      & 48.99 & 2.64  & 10.87 & 21.54             & \firstcolor 0.30              & \firstcolor 0.30  \\

			\hline
			
			Mean            & 8.40  & 9.89      & 24.66 & 28.30 & 17.20 & 15.94             & \secondcolor 5.82              & \firstcolor 5.55  \\
			STD             & 8.90  & 10.04     & 11.96 & 13.45 & 12.87 & 10.10             & 7.15              & 6.51  \\
			Median          & 4.30  & 5.20      & 24.53 & 29.40 & 17.78 & 17.96             & \firstcolor 2.07              & \secondcolor 2.90 \\
			
			\hline
		\end{tabu}
		
		\caption{Misclassification error ($\%$) for the estimation of multiple homographies.}
		\label{tab:adelaide_homography}		
	\end{subtable}
	
	\caption{Results on the AdelaideRMF dataset~\cite{wong11}. RPA/FABIA do not automatically determine the number of models (they draw this number from the ground truth). Grayed cells mark the first (dark gray) and second (light gray) best results among the  methods that automatically estimate this number.}
	\label{tab:adelaide}	
\end{table*}

We analyze in \cref{fig:adelaide_sigma} the performance of RS-NMU as $\sigma$ changes. We can see that the performance is not dramatically affected, providing to the user assurance that there is no need to fine-tune $\sigma$ in order to obtain good results.

In \cref{fig:adelaide_bad}, we present two image pairs for which RS-NMU presents sub-optimal results. We show with a simple argument that these results are not due to the NMU approach nor to the proposed NMU-ADMM. The group of green points in the top-left ground truth pair has errors, as pointed out in~\cite{Tepper2016arse}. This means that its effective size is smaller than what is marked in the ground truth. Thus, it does not pass the statistical test in \cref{sec:testing}. If we disable testing in our pipeline, we recover the error-less part of this group. However, disabling the testing might lead to recover other spurious groups as well, see the bottom row of \cref{fig:adelaide_bad}.

%
%

\section{Conclusions}
\label{sec:conclusions}

In this work, we first presented a highly efficient algorithm to address the nonnegative matrix underapproximation (NMU) problem. The solutions provided by our algorithm are the only ones in the literature for which the underapproximation constraint holds. NMU's results are interesting as, compared to traditional NMF, they present additional sparsity and part-based behavior without explicitly adding sparsity-inducing priors. To show this, we have presented a practical application to the analysis of climate data.

We have also presented an application to the task of robustly fitting multiple parametric models to a dataset. Accompanied by a specially designed algorithmic pipeline, NMU delivers state-of-the-art results, outperforming other alternatives in the literature.

In the future, we look forward to applying our NMU algorithm to the analysis of other datasets, since it will provide alternative insights to the ones obtained with other matrix factorization tools.

Finally, to obtain a completely autonomous and parameterless robust fitting system, we need to automatically estimate $\sigma$, not only for each image pair, but for each particular model, as different models might have different noise levels. An interesting lead would be to use the statistical test in \cref{sec:testing} as a way to infer the optimal $\sigma$ by computating the model's optimal p-value.

\clearpage

\balance

{\small

}

\end{document}